\PassOptionsToPackage{table}{xcolor}
\documentclass[10pt,twocolumn]{article}
\usepackage[preprint]{antintlpaper}

\usepackage{amsmath,amsfonts,bm}

\def\eqref#1{equation~\ref{#1}}

\def\1{\bm{1}}

\DeclareMathAlphabet{\mathsfit}{\encodingdefault}{\sfdefault}{m}{sl}
\SetMathAlphabet{\mathsfit}{bold}{\encodingdefault}{\sfdefault}{bx}{n}

\usepackage{float,algorithm,algorithmic,placeins}

\newcommand{\stitle}[1]{\par\noindent\textbf{#1}\enspace}
\newcommand{\chone}{Challenge~1}
\newcommand{\chtwo}{Challenge~2}
\newcommand{\chthree}{Challenge~3}
\graphicspath{{figures/}}

\AntTitle{Lightweight Tool Probing: Showing LLM Agents What Their Tools Return}
\AntAuthors{Yuchen~Mao\AntEqualContributor \and
  Zhihao~Wen\AntEqualContributor \and
  Yiwen~Zhao\AntEqualContributor \and
  Sijie~Dong\texorpdfstring{\textsuperscript{\dag}}{} \and
  Yuanyuan~Shi \and
  Yimin~Wang \texorpdfstring{\\}{, }
  Pan~Wang \and
  Hang~Wang \and
  Yitong~Lin \and
  Yihao~Hu \and
  Jiang-Ming~Yang \and
  Wei~Wu\texorpdfstring{\textsuperscript{\dag}}{}}
\AntAffiliations{Ant International, Ant Group}
\AntFirstPageNote{\textsuperscript{*}\,Equal Contributions\quad
  \textsuperscript{\dag}\,Corresponding Authors}
\AntDate{November 17, 2026}
\AntAbstract{%
Tool documentation tells an agent how to call a tool, but only calling it shows what it returns. Multi-turn function calling runs on exactly this missing knowledge. Arguments must match the formats a tool expects, and identifiers returned by one call must feed the next. Existing remedies either keep the agent reading, rewriting the documentation into better text about the tool, or keep it paying, rediscovering tool behavior inside every task; and task success alone cannot show whether such guidance helps. We introduce Lightweight Tool Probing (LTP): before any task, a model probes a toolset once in an isolated sandbox and keeps the raw responses it observed in a frozen evidence bank; during a task, it retrieves at most three short response excerpts per step, while identifiers still come from the live environment. LTP needs no gold solutions, no task feedback, and no parameter updates. To locate where the evidence helps, we also introduce execution diagnostics that score how much of a task's requirements an execution covers. 
On BFCL, LTP lifts DeepSeek-V4-Flash from 41.75\% to 49.75\% at 1.09 times the tokens of plain tool calling, outperforming all five baselines, the strongest of which need at least twice the tokens. The gain travels to GLM-5.3 and Qwen3.5-397B, and to ACEBench, AppWorld, and WorkBench. It is interpretable. The tasks LTP rescues gain argument-constraint coverage. Even its trajectories are better training data. They beat plain trajectories for all four student models, by up to 10.75 points.
\textbf{Showing an agent what its tools actually return is a cheap complement to telling it how to call them.}
}

\begin{document}
\raggedbottom
\makeanttitle

\section{Introduction}
\begin{figure*}[t]
    \centering
    \includegraphics[width=1\linewidth]{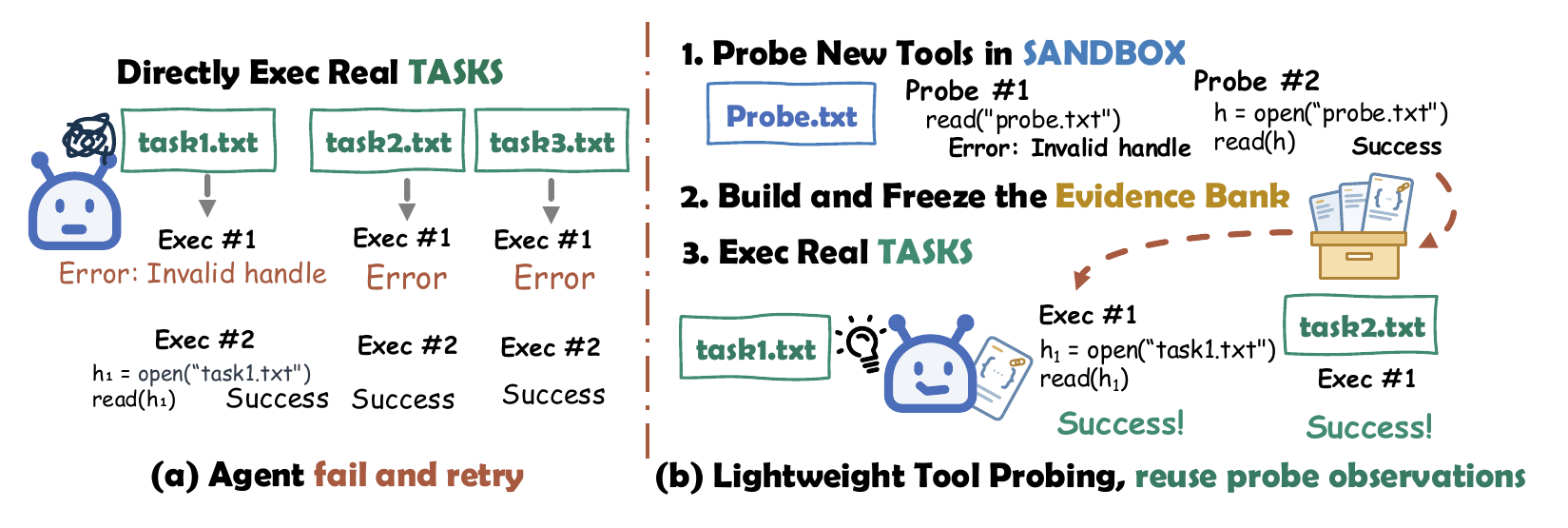}
    \vspace{-1.0em}
    \caption{\textbf{One sandbox probe replaces repeated in-task failures.} (a) Executing tasks directly, the agent calls \texttt{read} on an unopened file, fails, and retries the mistake in every new task. (b) With LTP, the same checkpoint probes the tools once, banks the observation, and succeeds on the first attempt after obtaining its own fresh handle. Illustrative file API: the call sequence is explanatory, not measured, and errors are never delivered as evidence.}
    \label{fig:probe-example}
    \vspace{-1.0em}
\end{figure*}

Language agents increasingly solve tasks by calling tools over many turns. Booking a trip, managing files, or operating a workplace application means chaining calls whose arguments depend on earlier results \citep{patil2025bfcl,yao2024taubench,trivedi2024appworld}. Tool documentation specifies how to call a tool but rarely shows what it returns, yet the return is exactly what the next call depends on, such as the formats to match, the units to convert, and the identifiers to carry from one call's output into the next call's arguments. Suppose an agent must read a file through an API it has never used (Figure~\ref{fig:probe-example}). The documentation lists \texttt{open(path)} and \texttt{read(handle)} but not what a handle looks like, so the agent calls \texttt{read} on the file name, fails, retries, and can repeat the mistake on every new task (Figure~\ref{fig:probe-example}a). Given the same documentation, a model can instead choose what to test, probe it in a sandbox before any task, and record what it observes: that \texttt{read} consumes the handle returned by \texttt{open}. Every later task can then retrieve that observation (Figure~\ref{fig:probe-example}b).

Existing remedies fail to let the agent observe a tool once and reuse what it saw without calling the tool again. Documentation rewriting refines tool descriptions \citep{yuan2025easytool,qu2025draft,fang2025play2prompt}, but agents must still translate this guidance into valid arguments and bind identifiers across calls. Errors in these steps can cause task failure even when the documentation is accurate. In-task feedback obtains real responses but pays for them inside every task and keeps nothing for the next one \citep{madaan2023selfrefine,shinn2023reflexion}, and RSIAgent builds a memory by exploration but verifies calls on every task \citep{zhu2026rsiagent}; in our experiments these remedies cost 1.7 to 2.3 times the tokens of plain tool calling. Training on interaction trajectories \citep{zeng2024agenttuning,zelikman2022star} moves the knowledge into the weights but needs gold solutions or a success signal. What is missing is a way to expose agents to actual tool outputs rather than rewritten descriptions, with preparation costs incurred once per toolset rather than per task, and independent of labels and training.
Studying and improving how agents acquire and use this knowledge raises three challenges.
\textbf{Challenge 1 (locating the failure).} Task success alone cannot tell whether an agent never investigated a tool, investigated it without obtaining usable evidence, or had the evidence and failed to apply it, and agents often discover useful information yet ignore it \citep{englander2026explore}. \textbf{Challenge 2 (transferring evidence across states at low cost).} Tool behavior becomes visible only when a tool is called, yet a response observed in one environment state does not guarantee that the same call succeeds in another task's state. Gathering that evidence must not repeat the work in every task. \textbf{Challenge 3 (isolating training transfer).} We must test whether probe-guided trajectories improve SFT outcomes when training tasks are held fixed and no evidence is retrieved at test time.

To address these challenges, we propose Lightweight Tool Probing (LTP). The model finds out how a toolset behaves by probing it once in an isolated sandbox before any task, and keeps the raw responses it observed in a frozen evidence bank, each linked to the call that produced it. During a task, LTP retrieves at most three response excerpts relevant to the current step, within 1,800 characters and marked as local to the sandbox, while the agent still obtains its own identifiers from the live environment (for \chtwo). It needs no gold solutions, no task success signal, and no parameter updates. 
Probing pays for itself. On BFCL \citep{patil2025bfcl} with DeepSeek-V4-Flash, \textbf{LTP gains 8.0 points at 1.09 times the tokens of plain tool calling}, beating all five baselines while adding the fewest tokens (Figure~\ref{fig:inference-tradeoff}). The gain is not tied to one model or benchmark. LTP also improves performance with GLM-5.3 and Qwen3.5-397B, as well as on ACEBench, AppWorld, and WorkBench. Execution diagnostics show that argument-constraint coverage increases on tasks LTP rescues and decreases on those it breaks (Section~\ref{sec:metrics-results}). The observations even outlive inference. Using LTP-generated trajectories for supervised fine-tuning (SFT) improves performance across all four student models, including a double-digit gain for Qwen3-14B.

Additional evaluations with a new bank confirm a 6.29-point gain over Plain on BFCL across three paired seeds. On 400 tasks written after submission and sealed before any run, DeepSeek-V4-Flash and Qwen3-8B gain 6.17 and 6.50 points. Content controls, eleven design ablations, and an evidence-removal SFT study further examine which observations and training signals carry these gains (Sections~\ref{sec:or-repeats}--\ref{sec:or-baselines}; Appendix~\ref{app:or-additional}).

Our contributions can be summarized as follows:
\begin{itemize}[topsep=-1pt,itemsep=1.5pt,parsep=0pt,partopsep=0pt,leftmargin=*]
    \item \textbf{Diagnostics that say where evidence helps, not only whether a task passed.} Three conditional gaps name the ways a tool goes unused, from never investigating it, to investigating without usable evidence, to having evidence but not applying it. Three coverage scores grade each executed trajectory against the task's requirements; they track success within each of six models and move with the outcomes LTP changes (Sections~\ref{sec:measurement}, \ref{sec:metrics-results}; \chone).
    \item \textbf{A cheap way to show an agent what its tools return.} Lightweight Tool Probing (LTP) lets a model try its tools once in a sandbox, before any task, and keep the raw responses it observed. Every later task reuses those observations at a fixed cost per step, with no labels, no task feedback, and no training. On BFCL, it beats five baselines at the lowest overhead (Sections~\ref{sec:ltp}, \ref{sec:inference}; \chtwo).
    \item \textbf{Evidence that also transfers through training, tested fairly.} Fine-tuned on LTP trajectories from the same teacher and tasks, all four student models beat their plain-trajectory controls with no bank at test time. For Qwen3-8B, LTP trajectories also beat trajectories generated by the ToolGrad and ToolACE-MT pipelines from the same teacher (Section~\ref{sec:sft-results}; \chthree).
\end{itemize}

\section{Related work}
\label{sec:related}
\stitle{Learning and evaluating tool use.}
Tool calling is learned from generated or curated call data \citep{schick2023toolformer,qin2023toolllm,patil2024gorilla,tang2023toolalpaca,liu2024apigen,liu2025toolace}, and agents are fine-tuned on interaction trajectories \citep{zeng2024agenttuning,chen2023fireact,chen2024agentflan,xi2024agentgym,zhang2025xlam,wang2024codeact}, synthetic instructions \citep{wang2023selfinstruct}, correctness-filtered self-generated solutions \citep{zelikman2022star,singh2024restem}, success/failure trajectory preferences \citep{song2024eto}, or experience from interaction and search \citep{su2025learnbyinteract,yu2025exact,li2026entropytools,zhao2026nextturn}; surveys map this space \citep{qin2024toollearning,qu2025toolsurvey}. Our SFT study compares jointly successful trajectories with and without probe evidence. Benchmarks cover call correctness \citep{li2023apibank,zhuang2023toolqa,huang2024metatool}, step-level utilization \citep{chen2024teval}, stable API evaluation \citep{guo2024stabletoolbench}, stateful multi-turn calling \citep{lu2025toolsandbox,patil2025bfcl,yao2024taubench}, and interactive environments \citep{liu2024agentbench,zhou2024webarena,farn2023tooltalk,trivedi2024appworld}. Most score outcomes; our diagnostics check whether an execution obtained, constrained, and bound the evidence its requirements demand, a verifier-free analogue of process supervision \citep{lightman2024verify}.

\stitle{Tool documentation and memory.}
Documentation alone enables zero-shot tool use \citep{hsieh2023tooldoc}; EasyTool condenses it into instructions \citep{yuan2025easytool}, and DRAFT and Play2Prompt refine it through tool interaction \citep{qu2025draft,fang2025play2prompt}. Other work distills reusable experience into skills and insights \citep{wang2023voyager,zhao2024expel}, workflows \citep{wang2024awm}, episodic memory \citep{park2023generativeagents}, or tool-side memory of capabilities and failure boundaries \citep{fang2026toolatlas}. Both lines hand the agent rewritten descriptions of tool behavior. Like DRAFT, Play2Prompt, and ToolAtlas, LTP probes reusably, but hands it the outputs themselves: raw, trace-linked response excerpts from the model's own bank, retrieved under a fixed budget \citep{lewis2020rag}.

The comparison with tool exploration can be made along three concrete dimensions: preparation inputs, stored content, and delivery timing. Play2Prompt synthesizes queries and uses LLM scoring to optimize documentation and selected examples; ToolAtlas uses synthesized tasks and LLM verifiers to construct graph memory and supplies guidance per task. LTP requires tool access, preserves calls with observed responses, and delivers up to three observations at each decision. The additional goal-free probing and representation experiments examine whether its gains require elaborate probe goals or a rewritten bank (Section~\ref{sec:or-content}).

\stitle{Feedback and exploration at inference.}
Self-Refine, Reflexion, CRITIC, and tree search revise or expand ReAct-style trajectories \citep{yao2023react} with in-task feedback \citep{madaan2023selfrefine,shinn2023reflexion,gou2024critic,yao2023tot,zhou2023lats}, though unaided self-correction has limits \citep{huang2024selfcorrect}. Simulators vet risky actions or refine calls before execution \citep{ruan2024toolemu,zhang2026gecko}, and RSIAgent explores a new environment to build memory without training \citep{zhu2026rsiagent}, but still verifies calls inside every task. Curiosity-driven and archive-based exploration \citep{pathak2017curiosity,ecoffet2021goexplore} and the finding that agents discover but ignore injected solutions \citep{englander2026explore} motivate separating investigation, evidence, and application. LTP instead pays once, before the first task, from public probe responses, and adds no revision, simulation, or verifier calls.

\section{Measuring the exploration-application gap}
\label{sec:measurement}

\subsection{Three conditional gaps}
\chone{} is to tell three failures apart: the agent never investigated a tool, investigated it without obtaining usable evidence, or had the evidence and failed to apply it. We make the three cases measurable through observable event chains rather than latent knowledge. Each chain tracks a tool condition independently specified in the annotation protocol. Before any task, $D$ records \emph{exposure} to the tool's interface or documentation. $I$ records a relevant \emph{investigation}, and $Q$ records \emph{qualified public probe evidence} supporting that condition. In the file example of Figure~\ref{fig:probe-example}, seeing \texttt{read} in the documentation is exposure, calling it in the sandbox is an investigation, and the recorded observation that \texttt{read} consumes the handle returned by \texttt{open} is qualified evidence. At task time, $O$ denotes an applicable execution \emph{opportunity}. $T$ means \emph{delivery} of the evidence before the relevant action, and $U$ represents \emph{correct use} under the registered condition. The three gaps are the conditional failure rates along this chain:
\begin{align}
G_1 &= 1-\Pr(I=1\mid D=1), \label{eq:g1}\\
G_2 &= 1-\Pr(Q=1\mid I=1), \label{eq:g2}\\
G_3 &= 1-\Pr(U=1\mid Q=1,O=1). \label{eq:g3}
\end{align}
In words, $G_1$ is how often exposure prompts no investigation. $G_2$ is how often investigation yields no usable evidence, and $G_3$ is how often evidence with an applicable opportunity is still not used correctly. The unit of analysis for $G_1$ and $G_2$ is a condition paired with a probing checkpoint; for $G_3$ it is a task-execution opportunity for that condition. Evidence can be available without being delivered, and an agent can act correctly without receiving it, so $G_3$ alone does not identify the causal effect of retrieved evidence. These gaps frame the execution diagnostics defined below, which Section~\ref{sec:metrics-results} evaluates; the supplementary material specifies the annotation protocol.

\subsection{Execution diagnostics}
\label{sec:diagnostics}
\begin{figure}[t]
\centering
\setlength{\abovecaptionskip}{5pt}

\includegraphics[width=\linewidth]{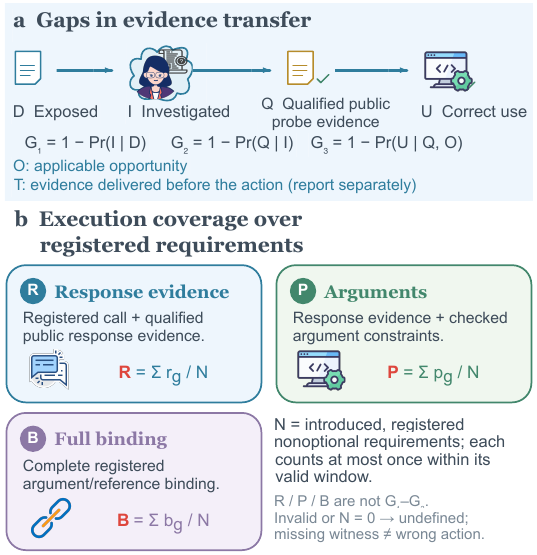}
\caption{\textbf{Where tool use fails, and how we score an execution.} (a) Three conditional gaps: after exposure ($D$), an agent may never investigate a tool ($I$), investigate without obtaining usable evidence ($Q$), or obtain evidence and still not use it correctly ($U$). (b) Three nested coverage scores of an executed trajectory, $B\le P\le R$: responses obtained, argument constraints met, and references bound. The scores grade executions and do not estimate the gaps; both panels are schematic.}
\label{fig:gaps}

\end{figure}
To locate where evidence helps inside an execution, we score how much of a task's requirements it covers, at three levels of strictness. The requirements are annotated once, before any experiment: each task's mandatory conditions, with their valid call paths and argument constraints. $\mathcal G_\tau$ collects the requirements that have already appeared in the user dialogue of trajectory $\tau$, so a requirement counts only after the user has asked for it, and $N=|\mathcal G_\tau|$. Each requirement $g$ then passes or fails three nested tests, each stricter than the last. \emph{Response} ($r_g=1$) means some annotated valid call path for $g$ received a qualified (nonempty, non-error) response. \emph{Argument} ($p_g=1$) means that call also satisfies its directly checkable argument constraints. \emph{Binding} ($b_g=1$) means every annotated argument is additionally supported by the trace, as a fixed value or a reference to an earlier output, such as a handle returned by \texttt{open}. Averaging over $\mathcal G_\tau$ gives response, argument-constraint, and binding coverage:
\begin{equation}
\label{eq:rpb}
\begin{aligned}
R(\tau)&=\tfrac{1}{N}\textstyle\sum_{g\in\mathcal G_\tau} r_g,\\
P(\tau)&=\tfrac{1}{N}\textstyle\sum_{g\in\mathcal G_\tau} p_g,\\
B(\tau)&=\tfrac{1}{N}\textstyle\sum_{g\in\mathcal G_\tau} b_g.
\end{aligned}
\end{equation}
Each requirement counts at most once within its valid time window, so $B\le P\le R$, and the scores are defined only for valid executions with $N>0$. Because they credit partial progress, they complement accuracy rather than replace it (Figure~\ref{fig:gaps}b).

\section{Lightweight Tool Probing}
\label{sec:ltp}
\begin{figure*}[t]
\centering
\includegraphics[width=\linewidth]{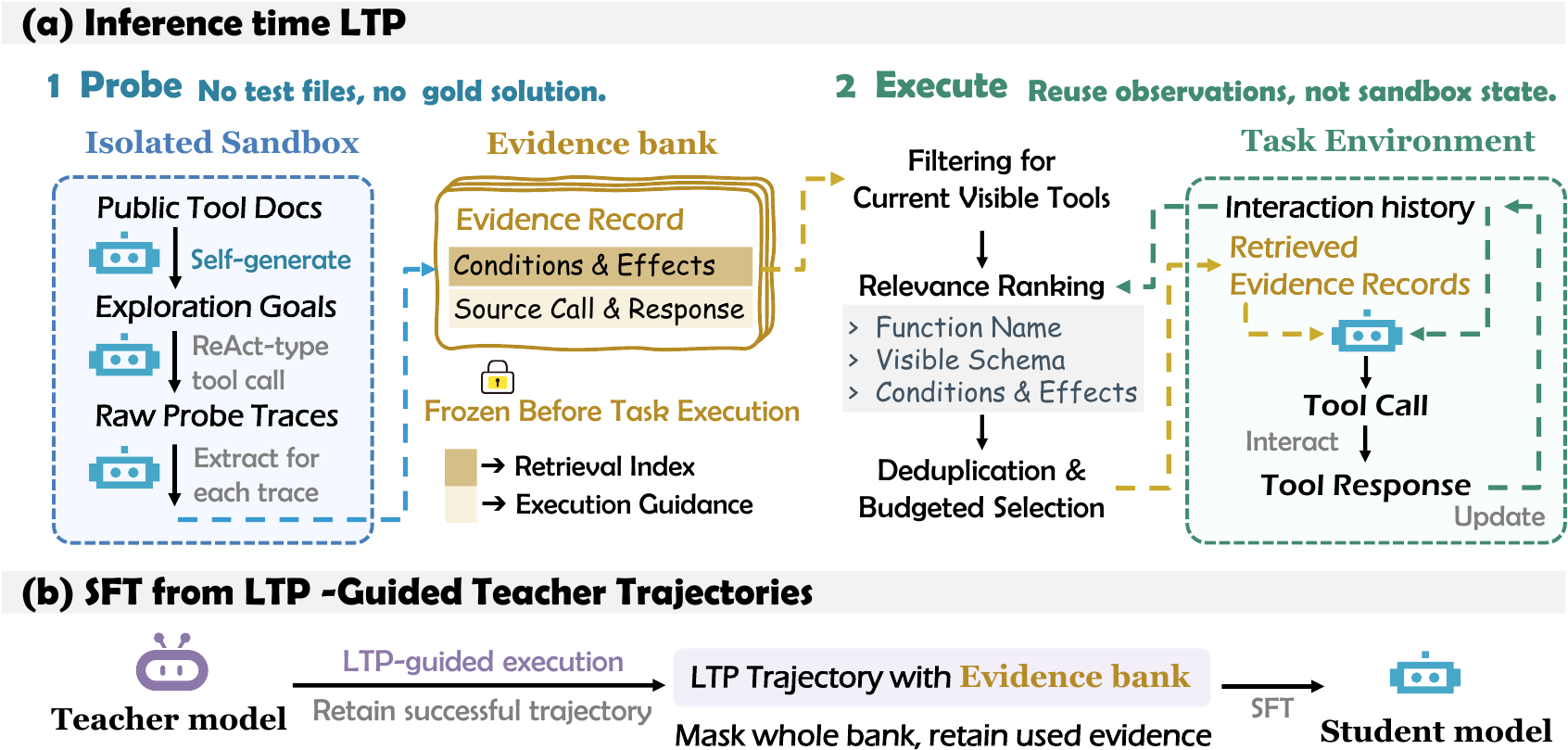}

\caption{\textbf{LTP probes once and retrieves raw responses.} (a) Inference: the same fixed checkpoint (1)~explores the public tools once in an isolated sandbox, (2)~stores each observed response with its source call in a frozen evidence bank, (3)~retrieves at most three eligible records within 1,800 characters at each decision, and (4)~acts in the task environment, obtaining its own state-dependent identifiers there. (b) Paired SFT: the same teacher and Train400 tasks yield 127 jointly successful Plain/LTP trajectory pairs; students train on either set, with the retrieved evidence kept as loss-masked context in LTP trajectories, and every checkpoint is evaluated with Plain inference and with its own bank.}
\label{fig:ltp-method}

\end{figure*}

LTP is a fixed protocol for prompting, evidence construction, and retrieval. It trains no probing policy and updates no parameters. It assumes public tool descriptions and schemas, observable tool responses, and an isolated, resettable sandbox. For a model $\theta$, task $x$, visible interaction history $h_t$, and sandbox $E_s$, LTP computes
\begin{align}
\mathcal B_\theta &= \operatorname{Build}\big(\operatorname{Probe}_\theta(E_s)\big), \label{eq:bank}\\
\begin{split}
e_t &= \operatorname{Retrieve}\big(\mathcal B_\theta, h_t, \text{visible schemas}\big),\\
a_t &\sim p_\theta(\cdot\mid x,h_t,e_t).
\end{split}\label{eq:retrieve}
\end{align}
The bank $\mathcal B_\theta$ is built once, before any task; the evidence $e_t$ is recomputed at each decision $t$ and conditions the next action $a_t$. Probing and execution share one fixed $\theta$, and the intervention is purely in-context: Plain is the special case $e_t\equiv\varnothing$. Figure~\ref{fig:ltp-method}(a) shows the four stages.

\stitle{(1) Probing once.}
The model proposes exploration goals from the public documentation, covering prerequisites, state changes, errors, and recovery, and pursues them in a fixed number of probe episodes (80 on BFCL) under fixed interaction budgets. Each episode starts from a fresh, task-independent sandbox state, and probing sees only public tool responses, never gold solutions, evaluator state, or benchmark test files.

\stitle{(2) A bank of raw observations.}
Each record of $\mathcal B_\theta$ is a tuple $\rho=(f,c,r,\delta)$: a tool $f$, a call $c$ issued during probing, the raw response $r$ it returned, and a short description $\delta$ of the observed conditions and effects, written by the same model. $\delta$ serves only as a retrieval index; what LTP delivers is $r$ itself: the details that decide an argument, field names, value formats, identifier shapes, are exactly what a summary can drop (for \chtwo). The extraction prompt keeps negative observations and uncertainty. The bank is frozen before any task and shared by all tasks.

\stitle{(3) Budgeted retrieval.}
At each decision $t$, LTP scores the records of the currently visible tools. A record is \emph{eligible} if $r$ is a nonempty, non-error JSON object or array and the arguments of $c$ pass the visible schema checks; error records stay in the bank but are never delivered, so that no excerpt demonstrates a failed call. Eligible records are ranked by lexical overlap with the user and tool messages in $h_t$, comparing the name of $f$, then its schema, then $\delta$, and keeping one record per tool. $e_t$ is then the longest ranked prefix whose rendered text, with a caution that all values are local to the sandbox, fits $k$ records and $m$ characters; if nothing fits, $e_t=\varnothing$ and the step is exactly Plain. We use $k=3$ and $m=1{,}800$ throughout.

\stitle{(4) Execution with the task's own state.}
The agent then acts through the benchmark's tool-calling interface. A delivered $r$ transfers what is environment-invariant, the field names, formats, and prerequisite structure a response exposes; state-bound values, such as handles or booking numbers, are re-derived from the agent's own calls in the live task (for \chtwo).

\stitle{Cost and properties.}
Probing and bank construction run once per model and environment, so their cost is shared by all tasks; at task time, LTP adds at most $m$ characters per decision and no extra model or tool calls. On BFCL, preparation takes 8.18 million tokens, and the whole protocol uses 1.09 times the tokens of Plain (Section~\ref{sec:inference}; for \chtwo). Because the bank is frozen before the first task and retrieval is deterministic, the evidence never adapts to a task, and $e_t=\varnothing$ recovers Plain when the bank holds nothing relevant.

\section{Experiments}
\label{sec:results}
Three research questions follow the three challenges. RQ1 asks whether the execution diagnostics separate successful from failed executions and move when retrieval changes an outcome (Section~\ref{sec:metrics-results}). RQ2 asks whether LTP improves accuracy at low token cost and carries over to other models and environments (Section~\ref{sec:inference}). RQ3 asks with tasks and teacher held fixed, whether LTP trajectories make better SFT data (Section~\ref{sec:sft-results}). 

\subsection{Setup}
\label{sec:setup}
\stitle{Benchmark and metrics.}
We use the four multi-turn categories of BFCL \citep{patil2025bfcl} (base, missing function, missing parameter, and long context) in two 400-task panels, Train400 and Test400. Inference is evaluated on all 800 tasks; SFT trains on Train400 trajectories and is evaluated on Test400. Out-of-distribution evaluations use ACEBench Agent \citep{chen2025acebench}, AppWorld Test \citep{trivedi2024appworld}, and WorkBench \citep{styles2024workbench}. Accuracy is the official task success over every task. The diagnostics $R$, $P$, and $B$ use the requirement annotations of Section~\ref{sec:diagnostics}, held fixed across the reported analyses, and are defined only for valid executions, those the scorer can parse with complete logs. We summarize their association with success by the area under the ROC curve (AUROC), computed from valid Plain executions over all 800 tasks. Paired diagnostic changes use tasks valid under both Plain and LTP.

\stitle{Models and conditions.}
DeepSeek-V4-Flash (the 0731 version) is the main inference model and the SFT teacher. Students are Qwen2.5-7B, Qwen3-8B, Qwen3.5-9B, and Qwen3-14B \citep{yang2025qwen3}. Qwen3.5-397B-A17B and GLM-5.3\citep{glm2024chatglm} are used in the cross-model and diagnostic panels. \emph{Plain} calls the benchmark tools directly; \emph{LTP} adds the checkpoint's own bank from 80 probe episodes with $k=3$ and $m=1{,}800$ (Section~\ref{sec:ltp}). Baselines are BFCL adaptations of EasyTool, Play2Prompt, Reflexion, RSIAgent \citep{zhu2026rsiagent}, and Gecko/GATS \citep{zhang2026gecko}. Token cost counts the evaluated execution, its helper calls, and preparation amortized over the tasks that share it, reported as a ratio to Plain.
\stitle{Training.}
\label{sec:sft-protocol}
We keep the 127 teacher--task pairs from Train400 that succeed under both Plain and LTP generation (Figure~\ref{fig:ltp-method}b), so that both training arms share the same tasks and teacher (for \chthree). \emph{Plain SFT} uses the Plain trajectories; \emph{Retained SFT} uses the paired LTP trajectories and keeps the bank context as loss-masked input. All runs use global batch size 8 and an assistant-only loss. Table~\ref{tab:sft127} reports Qwen2.5-7B at learning rate $5\times10^{-6}$, epoch 1; Qwen3-8B at $2\times10^{-6}$, epoch 3; Qwen3.5-9B at $2\times10^{-6}$, epoch 2; and Qwen3-14B at $2\times10^{-6}$, epoch 6, with Plain and Retained SFT matched on learning rate, epoch, and the other training settings (Appendix~\ref{app:setup-details}). Every checkpoint is evaluated with Plain inference and with LTP using its own bank. Figure~\ref{fig:sft8b-baselines} adds two Qwen3-8B students trained on trajectories that the same teacher generates for the Train400 tasks through fixed-task adaptations of ToolGrad \citep{zhou2025toolgrad}, which builds tool-call chains by textual-gradient refinement, and ToolACE-MT \citep{zeng2025toolacemt}, which drafts a whole multi-turn dialogue and refines it by mask-and-fill. Both generation pipelines produce 127 trajectories that pass the official task checker. One ToolGrad trajectory has a serialized training sequence of 167,266 tokens, exceeding the configured 163,840-token limit. We exclude this entire trajectory without truncation, leaving 126 ToolGrad trajectories and 127 ToolACE-MT trajectories for SFT.

\subsection{Do the diagnostics capture execution quality? (RQ1)}
\label{sec:metrics-results}
\stitle{Coverage separates successful from failed executions within every model and moves with outcome changes.}
Under Plain inference, argument-constraint coverage $P$ separates successful from failed executions with AUROC 0.84 to 0.86 for Qwen2.5-7B, Qwen3-8B, and Qwen3-14B and 0.65 to 0.75 for Qwen3.5-9B, Qwen3.5-397B, and DeepSeek-V4-Flash; $R$ shows the same separation between these groups, and $B$ ranges from 0.64 to 0.76 (Table~\ref{tab:metrics-story}). AUROC ranks a model's own executions rather than comparing models, and it is lower for the stronger models, likely because they already satisfy most constraints (mean $P$ of 0.76 for Qwen3.5-9B and 0.79 for DeepSeek-V4-Flash, versus 0.64 for Qwen2.5-7B and 0.59 for Qwen3-8B), which leaves less variation between their successes and failures. The diagnostics also move with the outcome of each task. On tasks that LTP rescues, mean $\Delta R$, $\Delta P$, and $\Delta B$ are positive for every model, and on tasks it breaks they are negative (Figure~\ref{fig:metrics-transitions}).

A complementary check considers the 618 tasks in the original main run whose success outcome does not change between Plain and LTP. Mean argument-constraint coverage moves by at most 0.002 on these tasks. This extends the rescued/harmed analysis with the unchanged-outcome group (Appendix~\ref{app:or-diagnostics}).

\begin{table*}[!htb]
\centering\small
\caption{\textbf{Coverage separates successful from failed executions within every model.} Plain-inference AUROC on BFCL (Train400 + Test400) for predicting task success from response ($R$), argument-constraint ($P$), and binding ($B$) coverage; 0.5 is chance. Six models, one run each; sample sizes in Appendix~\ref{app:setup-details}, two further models in Appendix~\ref{app:v41}.}
\label{tab:metrics-story}
\setlength{\tabcolsep}{3.5pt}\renewcommand{\arraystretch}{1.08}
\begin{tabular}{@{}lrrrrrr@{}}
\toprule
Score & Qwen2.5-7B & Qwen3-8B & Qwen3.5-9B & Qwen3-14B & Qwen3.5-397B & DeepSeek-V4-Flash \\
\midrule
$R$ (response) & 0.827 & 0.846 & 0.745 & 0.846 & 0.711 & 0.590 \\
$P$ (argument) & 0.849 & 0.859 & 0.747 & 0.846 & 0.720 & 0.655 \\
$B$ (binding) & 0.711 & 0.736 & 0.686 & 0.763 & 0.684 & 0.642 \\
\bottomrule
\end{tabular}
\end{table*}

\begin{figure*}[t]
\centering
\input{figures/results/metrics_transitions}\hfill
\input{figures/results/inference_tradeoff}
\end{figure*}

\begin{table*}[t]
\centering
\caption{\textbf{LTP reaches the highest accuracy at the smallest token overhead.} DeepSeek-V4-Flash on the 800 BFCL multi-turn tasks (200 per category). Acc.: task success (\%); Tok.: execution, helper, and amortized preparation tokens relative to Plain; $R/P/B$: response, argument-constraint, and binding coverage.}
\label{tab:inference800}
\small
\setlength{\tabcolsep}{2pt}\renewcommand{\arraystretch}{1.12}
\begin{tabular}{@{}l*{13}{r}@{}}
\toprule
 & \multicolumn{2}{c}{Base} & \multicolumn{2}{c}{Missing func.} & \multicolumn{2}{c}{Missing param.} & \multicolumn{2}{c}{Long context} & \multicolumn{5}{c}{Overall} \\
\cmidrule(lr){2-3}\cmidrule(lr){4-5}\cmidrule(lr){6-7}\cmidrule(lr){8-9}\cmidrule(lr){10-14}
Method & Acc. & Tok. & Acc. & Tok. & Acc. & Tok. & Acc. & Tok. & Acc. & Tok. & $R$ & $P$ & $B$ \\
\midrule
Plain & 54.00 & $1.000$ & 31.50 & $1.000$ & 32.00 & $1.000$ & 49.50 & $1.000$ & 41.75 & $1.000$ & 0.843 & 0.788 & 0.594 \\
Reflexion & 44.00 & $2.203$ & 34.00 & $2.139$ & 30.50 & $2.138$ & 35.00 & $2.203$ & 35.88 & $2.173$ & 0.797 & 0.744 & 0.567 \\
RSIAgent & 57.50 & $\underline{1.593}$ & 39.50 & $\underline{1.658}$ & 29.50 & $\underline{1.775}$ & 51.00 & $1.823$ & 44.38 & $\underline{1.734}$ & \textbf{0.855} & \textbf{0.798} & \textbf{0.601} \\
Play2Prompt & 53.00 & $3.182$ & 33.50 & $2.321$ & \textbf{46.00} & $2.783$ & 52.00 & $1.644$ & 46.13 & $2.305$ & 0.795 & 0.770 & 0.566 \\
EasyTool & \underline{58.50} & $2.389$ & \textbf{43.50} & $2.204$ & 42.00 & $2.279$ & 49.50 & $\underline{1.590}$ & \underline{48.38} & $2.021$ & 0.832 & \textbf{0.798} & \underline{0.600} \\
Gecko/GATS & 54.50 & $6.533$ & 39.00 & $6.376$ & 41.00 & $6.763$ & \textbf{59.00} & $2.952$ & \underline{48.38} & $5.209$ & 0.835 & 0.789 & 0.591 \\
\textbf{LTP (ours)} & \textbf{61.50} & $\mathbf{1.118}$ & \underline{40.00} & $\mathbf{1.097}$ & \underline{44.50} & $\mathbf{1.091}$ & \underline{53.00} & $\mathbf{1.079}$ & \textbf{49.75} & $\mathbf{1.092}$ & \underline{0.848} & \underline{0.797} & 0.598 \\
\bottomrule
\end{tabular}
\par\smallskip\parbox{\linewidth}{\footnotesize Bold/underline: best/second value per column, Plain excluded from the token ranking. $R/P/B$ use the 795 tasks valid for all methods; accuracy uses all 800.}
\end{table*}

\begin{table*}[t]
    \centering
    \caption{\textbf{The gain carries over to other models and other environments.} (a) BFCL, all 800 tasks, for three models. (b) Three further environments with DeepSeek-V4-Flash, scored by each environment's own checker. Values are task success (\%); $\Delta$ is LTP minus Plain.}
    \label{tab:model-ood}
    \begin{minipage}[t]{0.49\linewidth}
    \centering\footnotesize
    (a) BFCL 800: across models\par\smallskip
    \setlength{\tabcolsep}{3pt}\renewcommand{\arraystretch}{1.13}
    \begin{tabular}{@{}lrrr@{}}
    \toprule
    Model & Plain & LTP & $\Delta$ \\
    \midrule
    DeepSeek-V4-Flash & 41.75 & 49.75 & +8.00 \\
    Qwen3.5-397B & 55.13 & 56.75 & +1.63 \\
    GLM-5.3 & 54.13 & 58.63 & +4.50 \\
    \bottomrule
    \end{tabular}
    \end{minipage}\hfill
    \begin{minipage}[t]{0.49\linewidth}
    \centering\footnotesize
    (b) OOD: across environments\par\smallskip
    \setlength{\tabcolsep}{3pt}\renewcommand{\arraystretch}{1.13}
    \begin{tabular}{@{}lrrrr@{}}
    \toprule
    Environment & $N$ & Plain & LTP & $\Delta$ \\
    \midrule
    ACE Agent & 49 & 24.49 & 30.61 & +6.12 \\
    AppWorld Test & 585 & 13.33 & 20.85 & +7.52 \\
    WorkBench & 210 & 74.29 & 79.52 & +5.24 \\
    \bottomrule
    \end{tabular}
    \end{minipage}
    \par\smallskip\parbox{\linewidth}{\footnotesize
    (a) GLM-5.3 runs with reasoning enabled at low effort. (b) ACE: 49 English Agent tasks (19 multi-step $+$ 30 multi-turn); AppWorld: all 585 official Test tasks; WorkBench: all 210 official multi-domain tasks.}
    \end{table*}

\subsection{Does LTP improve inference accuracy, and at what cost? (RQ2)}
\label{sec:inference}
\stitle{LTP gains 8.0 points with 9.2\% more tokens than Plain.}
LTP raises the solved tasks from Plain's 334 (41.75\%) to 398 (49.75\%), gaining 3.5 to 12.5 points in every category (Table~\ref{tab:inference800}). This costs 118.02 million tokens in total, 109.85 million for execution and 8.18 million for preparation, or 1.092 times Plain's 108.05 million. Every baseline trails LTP in accuracy while spending more: EasyTool and Gecko/GATS come closest at 48.38\% (Gecko/GATS: 387 tasks) but need 2.021 and 5.209 times Plain's tokens (Gecko/GATS: 562.80 million); Play2Prompt reaches 46.13\% at 2.305 times, RSIAgent 44.38\% at 1.734 times, and Reflexion 35.88\% at 2.173 times.

Table~\ref{tab:inference800} reports results on all BFCL tasks. Diagnostic means use the 795 valid tasks with defined scores for all seven methods. LTP reaches $(R,P,B)=(.848,.797,.598)$, compared with Plain's $(.843,.788,.594)$. LTP thus raises task accuracy and mean coverage over Plain at once, although RSIAgent and EasyTool post higher means on some diagnostics. At the task level, LTP rescues 121 of the 797 tasks with valid diagnostics under both conditions and breaks 58; the rescued tasks gain .165 in $P$ on average, and the broken tasks lose .225. Figure~\ref{fig:metrics-transitions} reports fixed Plain/LTP comparisons over the same 800-task suite.

On the full BFCL evaluation, LTP improves Qwen3.5-397B-A17B from 441 to 454 successful tasks (+1.63 percentage points) and GLM-5.3 from 433 to 469 (+4.50 points; Table~\ref{tab:model-ood}). Across environments, DeepSeek-V4-Flash-0731 gains 6.12 points on ACEBench Agent \citep{chen2025acebench}, 7.52 on AppWorld Test \citep{trivedi2024appworld}, and 5.24 on WorkBench \citep{styles2024workbench}. Appendix~\ref{app:metric-trace-cases} illustrates how diagnostic changes correspond to recorded tool calls, and Appendix~\ref{app:inference-trace-cases} presents three tasks on which LTP succeeds while the named alternatives fail.

\subsection{Repeated runs and newly written tasks}
\label{sec:or-repeats}
We conduct an additional batch of experiments with DeepSeek-V4-Flash and a newly built bank of 352 records from 80 probe episodes. These runs form a separate batch from Table~\ref{tab:inference800}; comparisons below use controls from the same batch. On BFCL800, three paired seeds improve mean accuracy from 37.04\% to 43.33\%, a gain of 6.29 percentage points with a 95\% task-family bootstrap interval of $[4.17,8.42]$. All three pairs improve, as do all twelve combinations of task category and seed.

We also evaluate 400 tasks written after submission and sealed before any run. On these tasks, DeepSeek-V4-Flash gains 6.17 points and Qwen3-8B gains 6.50 points; the Qwen3-8B bank is rebuilt for each seed. Three retained GLM-5.3 pairs on BFCL give a mean gain of 3.96 points. Table~\ref{tab:or-repeats} reports every pair and distinguishes the task-family intervals from GLM-5.3's seed-level interval.

\subsection{Which evidence and design choices carry the gain?}
\label{sec:or-content}
\stitle{Return content under a matched token budget.}
We hold the probing run, bank, and selector fixed and change only the delivered content. Each representation is cut to the length of the shortest one, token for token at every step, and uses a common plain format. Raw records score 3.58 points above the same calls without responses on DeepSeek-V4-Flash ($[1.38,6.00]$), and 4.1 points above them on Qwen3-8B ($[0.1,8.7]$). On DeepSeek-V4-Flash, calls alone and other tools' records differ from Plain by $-0.62$ points ($[-2.87,1.58]$) and $+1.21$ points ($[-0.87,3.33]$), respectively. These interventions support the contribution of relevant response content beyond additional context or calls alone.

Faithful summaries and usage instructions preserve the same observations and yield similar accuracy to raw records: the matched-seed differences, Raw minus the rewrite, are $-0.17$ points ($[-2.04,1.71]$) and $+0.25$ points ($[-2.38,2.81]$). Both rewrites require a second model pass over the bank. Raw records provide the observations directly without that rewriting pass. Short notes written during probing are a different representation: on Qwen3-8B they score 35.67\%, 7.3 points below raw records ($[2.3,14.2]$). Appendix~\ref{app:or-content} gives the full representation table and distinguishes the token-matched renderer from LTP's deployed format and budget.

\stitle{Tool coverage and a small retrieval budget.}
Eleven variants on BFCL800 use two paired seeds, for which Plain scores 36.94\% and full LTP 44.31\%. Removing evidence for half the functions reduces accuracy by 2.75 points; retaining one record per function reduces it by 2.94 points. With 40 rather than 80 probe episodes, distributing the budget across all eight domains scores 41.00\%, compared with 38.00\% when it covers four domains. The difference is 3.00 points ($[0.38,5.63]$), indicating the importance of broad coverage.

Random function probing without proposed goals retains 83\% of the gain over Plain. Randomizing retrieval order reduces accuracy by 2.63 points. Reducing delivery to one record or 600 characters reduces it by 3.00 or 5.06 points, while increasing it to five records or 3,600 characters provides no improvement in the point estimate. The tested results favor broad probing and a modest, ranked evidence budget. All eleven variants and their intervals appear in Table~\ref{tab:or-ablations}.

\subsection{Additional baselines, reuse costs, and state variation}
\label{sec:or-baselines}
\stitle{Accuracy and token cost against the strongest baselines.}
The original LTP accuracy differences from EasyTool and Gecko/GATS are not statistically significant (exact McNemar $p=.47$ and $.54$), while their token ratios to Plain are 1.09, 2.02, and 5.21, respectively. Two new paired seeds against EasyTool likewise give a 95\% interval of $[-3.08,3.86]$ points for LTP minus EasyTool. In these runs, EasyTool uses 126.5M and 127.4M execution tokens per 800 tasks, versus LTP's 84.1M and 84.5M plus 5.14M once for preparation, approximately 30\% fewer tokens with LTP.

On Qwen3-8B and the 400 new tasks, the token-matched LTP-record condition reaches 43.00\%, versus 36.42\% for Plain, 36.25\% for Play2Prompt, and 42.00\% for EasyTool over three paired seeds. LTP records exceed Play2Prompt by 6.75 points ($[2.00,12.17]$). A separate Qwen3-8B comparison on BFCL800 evaluates our paper-based implementation of ToolAtlas under an 8.18M-token preparation cap: ToolAtlas scores 11.56\%, Plain 13.56\%, and LTP 16.13\%. Appendix~\ref{app:or-baselines} provides the comparison scopes and intervals.

\stitle{Amortization and changing state.}
The new bank costs 5.14M tokens, compared with 8.18M for the original bank. Across the three new BFCL seeds, LTP uses 256.8M tokens including preparation for 2,400 task executions, versus Plain's 259.6M. Using the reported execution rates, preparation is recovered against EasyTool within 100 tasks and against Gecko/GATS from the 15th task. These are reuse-cost projections; Appendix~\ref{app:or-cost} states their accounting and scope.

To test sensitivity to sandbox identifiers, we rename every identifier in the bank and evaluate Qwen3-8B on 200 file-system tasks over two seeds. Accuracy changes by $-0.50$ points ($[-2.25,1.00]$). This result is consistent with transferring response structure while obtaining state-dependent values from the live task. The experiment changes identifiers; changes to tool behavior require updating the affected domain's bank.

\begin{table*}[!htb]
\centering
\begin{minipage}[t]{0.46\linewidth}
\centering\vspace{0pt}
\caption{\textbf{Training on LTP trajectories beats training on plain trajectories for every student.} Matched SFT on 127 paired Train400 trajectories, evaluated on Test400 (\% task success). P/L: Plain inference / LTP inference with the student's own bank; Ours: Retained SFT; $\Delta$: Plain-inference gain of ours over Plain SFT.}
\label{tab:sft127}
\scriptsize
\setlength{\tabcolsep}{1pt}\renewcommand{\arraystretch}{1.17}
\begin{tabular}{@{}lrrrr@{}}
\toprule
Student & Untrained P/L & Plain SFT P/L & Ours P/L & $\Delta$ \\
\midrule
Qwen2.5-7B & 11.50/14.00 & 22.75/22.50 & 30.00/25.75 & $+7.25$ \\
Qwen3-8B & 15.25/18.50 & 34.00/35.75 & 42.00/37.00 & $+8.00$ \\
Qwen3.5-9B & 28.75/29.50 & 48.50/46.75 & 52.75/49.75 & $+4.25$ \\
Qwen3-14B & 13.25/19.00 & 31.25/30.25 & 42.00/42.50 & $+10.75$ \\
\bottomrule
\end{tabular}
\end{minipage}\hfill
\begin{minipage}[t]{0.49\linewidth}
\centering\vspace{0pt}
\makeatletter\def\@captype{figure}\makeatother

\caption{\textbf{LTP trajectories beat trajectories from other generation pipelines.} Qwen3-8B students under Plain inference on Test400, task success (\%); Plain SFT, Retained SFT, and the untrained model are the Table~\ref{tab:sft127} entries.}
\label{fig:sft8b-baselines}
\includegraphics[width=\linewidth]{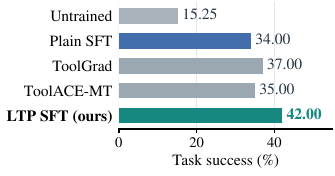}

\end{minipage}
\end{table*}

\begin{figure*}[t]
\centering
\includegraphics[width=\linewidth]{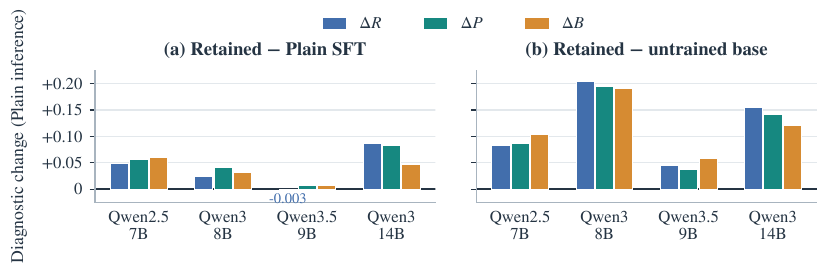}
\caption{\textbf{Retained SFT raises argument-constraint and binding coverage for every student.} Diagnostic changes of the Retained SFT checkpoints of Table~\ref{tab:sft127} under Plain inference, relative to (a) Plain SFT and (b) the untrained model, on each pair's jointly valid tasks.}
\label{fig:sft-diagnostics}
\end{figure*}

\subsection{Do LTP trajectories make better SFT data? (RQ3)}
\label{sec:sft-results}
\stitle{LTP trajectories can make better SFT data.}
Under Plain inference, the Retained SFT checkpoints beat their matched Plain SFT controls for all four students, by 4.25 to 10.75 points; the largest gap is Qwen3-14B, with 42.00\% versus 31.25\% (Table~\ref{tab:sft127}). The checkpoints also satisfy more argument constraints and bind more references than their controls. $P$ and $B$ rise for all four students and $R$ for three, with a 0.003 drop for Qwen3.5-9B (Figure~\ref{fig:sft-diagnostics}). LTP-generated trajectories can therefore improve task performance even when the student receives no bank at inference. At the same checkpoints, adding each student's own bank lowers accuracy by 3.00--5.00 points for three students and raises it by 0.50 points for Qwen3-14B (Table~\ref{tab:sft127}).

The smaller gains for Qwen3.5-9B may reflect a smaller remaining gap in tool-use knowledge. Its Plain SFT baseline
already achieves 48.50\% accuracy and .835 response coverage, leaving less additional benefit from LTP-generated
trajectories than for the other students. The small decrease in $R$ accompanies improvements in argument constraints,
bindings, and task success.

More broadly, the benefit of inference-time LTP depends on what the model already knows about the tools. SFT on LTP
trajectories may internalize part of the knowledge that retrieval otherwise supplies, reducing the additional
information provided by the bank. When this benefit becomes small, interference from the added context may outweigh
it. For example, on \texttt{multi\_turn\_miss\_func\_23}, the fine-tuned Qwen3.5-9B completes the task under Plain
inference, but repeatedly reads an empty file under LTP and fails to advance to the subsequent steps. At this
checkpoint, LTP rescues 22 tasks but turns 34 successes into failures, yielding a net accuracy decrease of 3.00
percentage points.

\stitle{Removing evidence from the training context has a small average effect.}
For Qwen2.5-7B, Qwen3-8B, and Qwen3.5-9B, we remove the retrieved evidence and compare 24 matched configurations spanning two learning rates and four epochs. Under Plain inference, evidence removal changes mean accuracy by +0.4 points: 13 configurations improve, 11 decline, and 21 of 24 intervals include zero. The teacher's action targets remain the same. This control supports the role of the LTP-generated action trajectories beyond the presence of loss-masked evidence in the context (Appendix~\ref{app:or-sft}).

\stitle{For Qwen3-8B, LTP trajectories also beat trajectories generated by other pipelines from the same teacher.}
Under Plain inference on Test400 (Figure~\ref{fig:sft8b-baselines}), Retained SFT reaches 42.00\%, against 34.00\% for Plain SFT, 37.00\% for ToolGrad, and 35.00\% for ToolACE-MT; the untrained model scores 15.25\%. The advantage occurs with the same teacher and comparable trajectory counts, so neither more data nor a stronger teacher explains it. Appendix~\ref{app:sft-trace-cases} shows three cases from the matched Plain versus Retained SFT comparison under Plain inference.

\section{Conclusion}
LTP starts from a simple asymmetry: documentation tells an agent how to call a tool, but only calling it shows what it returns. The protocol therefore probes a toolset once in an isolated sandbox, banks the raw responses rather than rewritten summaries, and retrieves at most a few short excerpts per decision, while identifiers and other task state still come from the live environment. The cost of learning how a toolset behaves is thus paid once per toolset, never inside a task, and the intervention stays purely in-context: no gold solutions, no task feedback, no parameter updates.
On BFCL, LTP raises DeepSeek-V4-Flash by 8.0 points at 1.09 times the tokens of plain tool calling, while the five baselines spend 1.7 to 5.2 times the tokens for less, and the gain carries over to two other models, and to three other environments. The execution diagnostics show that the gain is no accident of the checker: on average, the tasks LTP rescues satisfy more argument constraints and the tasks it breaks satisfy fewer. The observations even outlive inference: with tasks and teacher matched, checkpoints fine-tuned on LTP trajectories beat their matched controls for all four student models, without any bank at test time.
Together, these results argue for a division of labor: sandbox evidence carries what is invariant about a tool, the live task supplies what is not, and either a retrieval step or the training data can deliver the evidence. \textbf{Showing an agent what its tools actually return is a cheap complement to telling it how to call them.}

The additional studies reinforce this picture across paired runs and newly written tasks. They identify relevant response content, broad tool coverage, and a small ranked retrieval budget as useful ingredients. Faithful rewrites can preserve the same information, while raw records provide it without a second rewriting pass. The evidence-removal SFT control further supports transferring the benefit through the teacher's action trajectories.

\FloatBarrier
\section*{AI use statement}
We used large language models (LLMs) as assistants for experiment planning, code implementation, trajectory processing, result analysis, literature search, manuscript drafting, and figure generation. Separately from this assistance, LLMs are part of the experiments themselves: they are the evaluated agents, they generate the teacher trajectories used for fine-tuning, and they produced the initial requirement annotations behind the execution diagnostics. The authors reviewed all AI-assisted text, code, and figures, checked the reported results against the experiment records, and take full responsibility for the content of this paper.

\section*{Reproducibility statement}
Sections~\ref{sec:measurement} and~\ref{sec:ltp} define the diagnostics and LTP; Section~\ref{sec:setup} states the models, task panels, baselines, and training settings; Appendix~\ref{app:setup-details} adds benchmark handling, valid sample counts, cost accounting, and seeds. All benchmarks and models are publicly available. The supplementary material contains our code for probing, bank construction, retrieval, the diagnostics, the baseline adaptations, and SFT.

\bibliography{references}

\begin{thebibliography}{64}
\providecommand{\natexlab}[1]{#1}
\providecommand{\url}[1]{\texttt{#1}}
\expandafter\ifx\csname urlstyle\endcsname\relax
  \providecommand{\doi}[1]{doi: #1}\else
  \providecommand{\doi}{doi: \begingroup \urlstyle{rm}\Url}\fi

\bibitem[Chen et~al.(2023)Chen, Shu, Shareghi, Collier, Narasimhan, and
  Yao]{chen2023fireact}
Baian Chen, Chang Shu, Ehsan Shareghi, Nigel Collier, Karthik Narasimhan, and
  Shunyu Yao.
\newblock Fireact: Toward language agent fine-tuning.
\newblock \emph{arXiv preprint arXiv:2310.05915}, 2023.

\bibitem[Chen et~al.(2025)Chen, Hao, Liu, Huang, Zeng, Yu, Li, Huang, Liu,
  Wang, et~al.]{chen2025acebench}
Chen Chen, Xinlong Hao, Weiwen Liu, Xu~Huang, Xingshan Zeng, Shuai Yu, Dexun
  Li, Yuefeng Huang, Xiangcheng Liu, Xinzhi Wang, et~al.
\newblock Acebench: A comprehensive evaluation of llm tool usage.
\newblock In \emph{EMNLP (Findings)}, pages 12970--12998, 2025.

\bibitem[Chen et~al.(2024{\natexlab{a}})Chen, Du, Zhang, Liu, Liu, Zheng, Zhuo,
  Zhang, Lin, Chen, et~al.]{chen2024teval}
Zehui Chen, Weihua Du, Wenwei Zhang, Kuikun Liu, Jiangning Liu, Miao Zheng,
  Jingming Zhuo, Songyang Zhang, Dahua Lin, Kai Chen, et~al.
\newblock T-eval: Evaluating the tool utilization capability of large language
  models step by step.
\newblock In \emph{Proceedings of the 62nd Annual Meeting of the Association
  for Computational Linguistics (Volume 1: Long Papers)}, pages 9510--9529,
  2024{\natexlab{a}}.

\bibitem[Chen et~al.(2024{\natexlab{b}})Chen, Liu, Wang, Zhang, Liu, Lin, Chen,
  and Zhao]{chen2024agentflan}
Zehui Chen, Kuikun Liu, Qiuchen Wang, Wenwei Zhang, Jiangning Liu, Dahua Lin,
  Kai Chen, and Feng Zhao.
\newblock {Agent-FLAN}: Designing data and methods of effective agent tuning
  for large language models.
\newblock In \emph{Findings of the Association for Computational Linguistics:
  ACL 2024}, 2024{\natexlab{b}}.
\newblock URL \url{https://arxiv.org/abs/2403.12881}.

\bibitem[Ecoffet et~al.(2021)Ecoffet, Huizinga, Lehman, Stanley, and
  Clune]{ecoffet2021goexplore}
Adrien Ecoffet, Joost Huizinga, Joel Lehman, Kenneth~O. Stanley, and Jeff
  Clune.
\newblock First return, then explore.
\newblock \emph{Nature}, 590:\penalty0 580--586, 2021.
\newblock URL \url{https://arxiv.org/abs/2004.12919}.

\bibitem[Engl{\"a}nder et~al.(2026)Engl{\"a}nder, Althammer, {\"U}st{\"u}n,
  Gall{\'e}, and Sherborne]{englander2026explore}
Leon Engl{\"a}nder, Sophia Althammer, Ahmet {\"U}st{\"u}n, Matthias Gall{\'e},
  and Tom Sherborne.
\newblock Agents explore but agents ignore: Llms lack environmental curiosity.
\newblock \emph{arXiv preprint arXiv:2604.17609}, 2026.

\bibitem[Fang et~al.(2025)Fang, Zhang, Qian, Glass, and
  Zhu]{fang2025play2prompt}
Wei Fang, Yang Zhang, Kaizhi Qian, James Glass, and Yada Zhu.
\newblock Play2prompt: Zero-shot tool instruction optimization for llm agents
  via tool play.
\newblock In \emph{Findings of the Association for Computational Linguistics:
  ACL 2025}, pages 26274--26290, 2025.

\bibitem[Fang et~al.(2026)Fang, Yang, Yang, Qin, Li, Lin, Rajmohan, and
  Zhang]{fang2026toolatlas}
Yue Fang, Zhibang Yang, Fangkai Yang, Xiaoting Qin, Liqun Li, Qingwei Lin,
  Saravan Rajmohan, and Dongmei Zhang.
\newblock Toolatlas: Learning once, reusing everywhere with tool-side memory.
\newblock \emph{arXiv preprint arXiv:2607.11126}, 2026.

\bibitem[Farn and Shin(2023)]{farn2023tooltalk}
Nicholas Farn and Richard Shin.
\newblock Tooltalk: Evaluating tool-usage in a conversational setting.
\newblock \emph{arXiv preprint arXiv:2311.10775}, 2023.

\bibitem[Gou et~al.(2024)Gou, Shao, Gong, Shen, Yang, Duan, and
  Chen]{gou2024critic}
Zhibin Gou, Zhihong Shao, Yeyun Gong, Yelong Shen, Yujiu Yang, Nan Duan, and
  Weizhu Chen.
\newblock {CRITIC}: Large language models can self-correct with
  tool-interactive critiquing.
\newblock In \emph{International Conference on Learning Representations
  (ICLR)}, 2024.
\newblock URL \url{https://arxiv.org/abs/2305.11738}.

\bibitem[Guo et~al.(2024)Guo, Cheng, Wang, Liang, Qin, Li, Liu, Sun, and
  Liu]{guo2024stabletoolbench}
Zhicheng Guo, Sijie Cheng, Hao Wang, Shihao Liang, Yujia Qin, Peng Li, Zhiyuan
  Liu, Maosong Sun, and Yang Liu.
\newblock Stabletoolbench: Towards stable large-scale benchmarking on tool
  learning of large language models.
\newblock In \emph{Findings of the Association for Computational Linguistics:
  ACL 2024}, pages 11143--11156, 2024.

\bibitem[Hsieh et~al.(2023)Hsieh, Chen, Li, Fujii, Ratner, Lee, Krishna, and
  Pfister]{hsieh2023tooldoc}
Cheng-Yu Hsieh, Si-An Chen, Chun-Liang Li, Yasuhisa Fujii, Alexander Ratner,
  Chen-Yu Lee, Ranjay Krishna, and Tomas Pfister.
\newblock Tool documentation enables zero-shot tool-usage with large language
  models.
\newblock \emph{arXiv preprint arXiv:2308.00675}, 2023.

\bibitem[Huang et~al.(2024{\natexlab{a}})Huang, Chen, Mishra, Zheng, Yu, Song,
  and Zhou]{huang2024selfcorrect}
Jie Huang, Xinyun Chen, Swaroop Mishra, Huaixiu~Steven Zheng, Adams~Wei Yu,
  Xinying Song, and Denny Zhou.
\newblock Large language models cannot self-correct reasoning yet.
\newblock In \emph{International Conference on Learning Representations
  (ICLR)}, 2024{\natexlab{a}}.
\newblock URL \url{https://arxiv.org/abs/2310.01798}.

\bibitem[Huang et~al.(2024{\natexlab{b}})Huang, Shi, Li, Fan, Wu, Zhang, Liu,
  Zhou, Wan, Gong, et~al.]{huang2024metatool}
Yue Huang, Jiawen Shi, Yuan Li, Chenrui Fan, Siyuan Wu, Qihui Zhang, Yixin Liu,
  Pan Zhou, Yao Wan, Neil Gong, et~al.
\newblock Metatool benchmark for large language models: Deciding whether to use
  tools and which to use.
\newblock In \emph{International Conference on Learning Representations},
  volume 2024, pages 42978--43007, 2024{\natexlab{b}}.

\bibitem[Lewis et~al.(2020)Lewis, Perez, Piktus, Petroni, Karpukhin, Goyal,
  K{\"u}ttler, Lewis, Yih, Rockt{\"a}schel, Riedel, and Kiela]{lewis2020rag}
Patrick Lewis, Ethan Perez, Aleksandra Piktus, Fabio Petroni, Vladimir
  Karpukhin, Naman Goyal, Heinrich K{\"u}ttler, Mike Lewis, Wen-tau Yih, Tim
  Rockt{\"a}schel, Sebastian Riedel, and Douwe Kiela.
\newblock Retrieval-augmented generation for knowledge-intensive {NLP} tasks.
\newblock In \emph{Advances in Neural Information Processing Systems
  (NeurIPS)}, 2020.
\newblock URL \url{https://arxiv.org/abs/2005.11401}.

\bibitem[Li et~al.(2023)Li, Zhao, Yu, Song, Li, Yu, Li, Huang, and
  Li]{li2023apibank}
Minghao Li, Yingxiu Zhao, Bowen Yu, Feifan Song, Hangyu Li, Haiyang Yu, Zhoujun
  Li, Fei Huang, and Yongbin Li.
\newblock Api-bank: A comprehensive benchmark for tool-augmented llms.
\newblock In \emph{Proceedings of the 2023 conference on empirical methods in
  natural language processing}, pages 3102--3116, 2023.

\bibitem[Li et~al.(2026)Li, Wang, Zhao, Chen, Li, Chen, Cao, Ye, Chai, and
  Yin]{li2026entropytools}
Zeping Li, Hongru Wang, Yiwen Zhao, Guanhua Chen, Yixia Li, Keyang Chen, Yixin
  Cao, Guangnan Ye, Hongfeng Chai, and Zhenfei Yin.
\newblock Rethinking the role of entropy in optimizing tool-use behaviors for
  large language model agents.
\newblock In \emph{Proceedings of the 64th Annual Meeting of the Association
  for Computational Linguistics (Volume 1: Long Papers)}, pages 27955--27967,
  2026.

\bibitem[Lightman et~al.(2024)Lightman, Kosaraju, Burda, Edwards, Baker, Lee,
  Leike, Schulman, Sutskever, and Cobbe]{lightman2024verify}
Hunter Lightman, Vineet Kosaraju, Yura Burda, Harri Edwards, Bowen Baker, Teddy
  Lee, Jan Leike, John Schulman, Ilya Sutskever, and Karl Cobbe.
\newblock Let's verify step by step.
\newblock In \emph{International Conference on Learning Representations
  (ICLR)}, 2024.
\newblock URL \url{https://arxiv.org/abs/2305.20050}.

\bibitem[Liu et~al.(2025)Liu, Huang, Zeng, Yu, Li, Wang, Gan, Liu, Yu, WANG,
  et~al.]{liu2025toolace}
Weiwen Liu, Xu~Huang, Xingshan Zeng, Shuai Yu, Dexun Li, Shuai Wang, Weinan
  Gan, Zhengying Liu, Yuanqing Yu, Zezhong WANG, et~al.
\newblock Toolace: Winning the points of llm function calling.
\newblock In \emph{International conference on learning representations},
  volume 2025, pages 41359--41381, 2025.

\bibitem[Liu et~al.(2024{\natexlab{a}})Liu, Yu, Zhang, Xu, Lei, Lai, Gu, Ding,
  Men, Yang, et~al.]{liu2024agentbench}
Xiao Liu, Hao Yu, Hanchen Zhang, Yifan Xu, Xuanyu Lei, Hanyu Lai, Yu~Gu,
  Hangliang Ding, Kaiwen Men, Kejuan Yang, et~al.
\newblock Agentbench: Evaluating llms as agents.
\newblock In \emph{International Conference on Learning Representations},
  volume 2024, pages 52989--53046, 2024{\natexlab{a}}.

\bibitem[Liu et~al.(2024{\natexlab{b}})Liu, Hoang, Zhang, Zhu, Lan, Kokane,
  Tan, Yao, Liu, Feng, et~al.]{liu2024apigen}
Zuxin Liu, Thai Hoang, Jianguo Zhang, Ming Zhu, Tian Lan, Shirley Kokane,
  Juntao Tan, Weiran Yao, Zhiwei Liu, Yihao Feng, et~al.
\newblock Apigen: Automated pipeline for generating verifiable and diverse
  function-calling datasets.
\newblock \emph{Advances in Neural Information Processing Systems},
  37:\penalty0 54463--54482, 2024{\natexlab{b}}.

\bibitem[Lu et~al.(2025)Lu, Holleis, Zhang, Aumayer, Nan, Bai, Ma, Ma, Li, Yin,
  et~al.]{lu2025toolsandbox}
Jiarui Lu, Thomas Holleis, Yizhe Zhang, Bernhard Aumayer, Feng Nan, Haoping
  Bai, Shuang Ma, Shen Ma, Mengyu Li, Guoli Yin, et~al.
\newblock Toolsandbox: A stateful, conversational, interactive evaluation
  benchmark for llm tool use capabilities.
\newblock In \emph{Findings of the Association for Computational Linguistics:
  NAACL 2025}, pages 1160--1183, 2025.

\bibitem[Madaan et~al.(2023)Madaan, Tandon, Gupta, Hallinan, Gao, Wiegreffe,
  Alon, Dziri, Prabhumoye, Yang, et~al.]{madaan2023selfrefine}
Aman Madaan, Niket Tandon, Prakhar Gupta, Skyler Hallinan, Luyu Gao, Sarah
  Wiegreffe, Uri Alon, Nouha Dziri, Shrimai Prabhumoye, Yiming Yang, et~al.
\newblock Self-refine: Iterative refinement with self-feedback.
\newblock \emph{Advances in neural information processing systems},
  36:\penalty0 46534--46594, 2023.

\bibitem[Park et~al.(2023)Park, O'Brien, Cai, Morris, Liang, and
  Bernstein]{park2023generativeagents}
Joon~Sung Park, Joseph~C. O'Brien, Carrie~J. Cai, Meredith~Ringel Morris, Percy
  Liang, and Michael~S. Bernstein.
\newblock Generative agents: Interactive simulacra of human behavior.
\newblock In \emph{Proceedings of the 36th Annual ACM Symposium on User
  Interface Software and Technology (UIST)}, 2023.
\newblock URL \url{https://arxiv.org/abs/2304.03442}.

\bibitem[Pathak et~al.(2017)Pathak, Agrawal, Efros, and
  Darrell]{pathak2017curiosity}
Deepak Pathak, Pulkit Agrawal, Alexei~A. Efros, and Trevor Darrell.
\newblock Curiosity-driven exploration by self-supervised prediction.
\newblock In \emph{International Conference on Machine Learning (ICML)}, 2017.
\newblock URL \url{https://arxiv.org/abs/1705.05363}.

\bibitem[Patil et~al.(2024)Patil, Zhang, Wang, and Gonzalez]{patil2024gorilla}
Shishir~G Patil, Tianjun Zhang, Xin Wang, and Joseph~E Gonzalez.
\newblock Gorilla: Large language model connected with massive apis.
\newblock \emph{Advances in Neural Information Processing Systems},
  37:\penalty0 126544--126565, 2024.

\bibitem[Patil et~al.(2025)Patil, Mao, Yan, Ji, Suresh, Stoica, and
  Gonzalez]{patil2025bfcl}
Shishir~G Patil, Huanzhi Mao, Fanjia Yan, Charlie Cheng-Jie Ji, Vishnu Suresh,
  Ion Stoica, and Joseph~E Gonzalez.
\newblock The berkeley function calling leaderboard (bfcl): From tool use to
  agentic evaluation of large language models.
\newblock In \emph{Forty-second International Conference on Machine Learning},
  2025.

\bibitem[Qin et~al.(2024{\natexlab{a}})Qin, Hu, Lin, Chen, Ding, Cui, Zeng,
  Zhou, Huang, Xiao, et~al.]{qin2024toollearning}
Yujia Qin, Shengding Hu, Yankai Lin, Weize Chen, Ning Ding, Ganqu Cui, Zheni
  Zeng, Xuanhe Zhou, Yufei Huang, Chaojun Xiao, et~al.
\newblock Tool learning with foundation models.
\newblock \emph{ACM Computing Surveys}, 57\penalty0 (4):\penalty0 1--40,
  2024{\natexlab{a}}.

\bibitem[Qin et~al.(2024{\natexlab{b}})Qin, Liang, Ye, Zhu, Yan, Lu, Lin, Cong,
  Tang, Qian, et~al.]{qin2023toolllm}
Yujia Qin, Shihao Liang, Yining Ye, Kunlun Zhu, Lan Yan, Yaxi Lu, Yankai Lin,
  Xin Cong, Xiangru Tang, Bill Qian, et~al.
\newblock Toolllm: Facilitating large language models to master 16000+
  real-world apis.
\newblock In \emph{International Conference on Learning Representations},
  volume 2024, pages 9695--9717, 2024{\natexlab{b}}.

\bibitem[Qu et~al.(2025{\natexlab{a}})Qu, Dai, Wei, Cai, Wang, Yin, Xu, and
  Wen]{qu2025draft}
Changle Qu, Sunhao Dai, Xiaochi Wei, Hengyi Cai, Shuaiqiang Wang, Dawei Yin,
  Jun Xu, and Ji-Rong Wen.
\newblock From exploration to mastery: Enabling llms to master tools via
  self-driven interactions.
\newblock In \emph{International Conference on Learning Representations},
  volume 2025, pages 55675--55697, 2025{\natexlab{a}}.

\bibitem[Qu et~al.(2025{\natexlab{b}})Qu, Dai, Wei, Cai, Wang, Yin, Xu, and
  Wen]{qu2025toolsurvey}
Changle Qu, Sunhao Dai, Xiaochi Wei, Hengyi Cai, Shuaiqiang Wang, Dawei Yin,
  Jun Xu, and Ji-Rong Wen.
\newblock Tool learning with large language models: A survey.
\newblock \emph{Frontiers of Computer Science}, 19\penalty0 (8):\penalty0
  198343, 2025{\natexlab{b}}.

\bibitem[Ruan et~al.(2024)Ruan, Dong, Wang, Pitis, Zhou, Ba, Dubois, Maddison,
  and Hashimoto]{ruan2024toolemu}
Yangjun Ruan, Honghua Dong, Andrew Wang, Silviu Pitis, Yongchao Zhou, Jimmy Ba,
  Yann Dubois, Chris~J. Maddison, and Tatsunori Hashimoto.
\newblock Identifying the risks of {LM} agents with an {LM}-emulated sandbox.
\newblock In \emph{International Conference on Learning Representations
  (ICLR)}, 2024.
\newblock URL \url{https://arxiv.org/abs/2309.15817}.

\bibitem[Schick et~al.(2023)Schick, Dwivedi-Yu, Dess{\`\i}, Raileanu, Lomeli,
  Hambro, Zettlemoyer, Cancedda, and Scialom]{schick2023toolformer}
Timo Schick, Jane Dwivedi-Yu, Roberto Dess{\`\i}, Roberta Raileanu, Maria
  Lomeli, Eric Hambro, Luke Zettlemoyer, Nicola Cancedda, and Thomas Scialom.
\newblock Toolformer: Language models can teach themselves to use tools.
\newblock \emph{Advances in neural information processing systems},
  36:\penalty0 68539--68551, 2023.

\bibitem[Shinn et~al.(2023)Shinn, Cassano, Gopinath, Narasimhan, and
  Yao]{shinn2023reflexion}
Noah Shinn, Federico Cassano, Ashwin Gopinath, Karthik Narasimhan, and Shunyu
  Yao.
\newblock Reflexion: Language agents with verbal reinforcement learning.
\newblock \emph{Advances in neural information processing systems},
  36:\penalty0 8634--8652, 2023.

\bibitem[Singh et~al.(2024)Singh, Co-Reyes, Agarwal, Anand, Patil, Garcia, Liu,
  Harrison, Lee, Xu, Parisi, Kumar, Alemi, Rizkowsky, Nova, Adlam, Bohnet,
  Elsayed, Sedghi, Mordatch, Simpson, Gur, Snoek, Pennington, Hron, Kenealy,
  Swersky, Mahajan, Culp, Xiao, Bileschi, Constant, Novak, Liu, Warkentin,
  Qian, Bansal, Dyer, Neyshabur, Sohl-Dickstein, and Fiedel]{singh2024restem}
Avi Singh, John~D. Co-Reyes, Rishabh Agarwal, Ankesh Anand, Piyush Patil,
  Xavier Garcia, Peter~J. Liu, James Harrison, Jaehoon Lee, Kelvin Xu, Aaron
  Parisi, Abhishek Kumar, Alex Alemi, Alex Rizkowsky, Azade Nova, Ben Adlam,
  Bernd Bohnet, Gamaleldin Elsayed, Hanie Sedghi, Igor Mordatch, Isabelle
  Simpson, Izzeddin Gur, Jasper Snoek, Jeffrey Pennington, Jiri Hron, Kathleen
  Kenealy, Kevin Swersky, Kshiteej Mahajan, Laura Culp, Lechao Xiao, Maxwell~L.
  Bileschi, Noah Constant, Roman Novak, Rosanne Liu, Tris Warkentin, Yundi
  Qian, Yamini Bansal, Ethan Dyer, Behnam Neyshabur, Jascha Sohl-Dickstein, and
  Noah Fiedel.
\newblock Beyond human data: Scaling self-training for problem-solving with
  language models.
\newblock \emph{Transactions on Machine Learning Research}, 2024.
\newblock URL \url{https://arxiv.org/abs/2312.06585}.

\bibitem[Song et~al.(2024)Song, Yin, Yue, Huang, Li, and Lin]{song2024eto}
Yifan Song, Da~Yin, Xiang Yue, Jie Huang, Sujian Li, and Bill~Yuchen Lin.
\newblock Trial and error: Exploration-based trajectory optimization for {LLM}
  agents.
\newblock In \emph{Proceedings of ACL 2024 (Volume 1: Long Papers)}, 2024.
\newblock URL \url{https://arxiv.org/abs/2403.02502}.

\bibitem[Styles et~al.(2024)Styles, Miller, Cerda-Mardini, Guha, Sanchez, and
  Vidgen]{styles2024workbench}
Olly Styles, Sam Miller, Patricio Cerda-Mardini, Tanaya Guha, Victor Sanchez,
  and Bertie Vidgen.
\newblock {WorkBench}: a benchmark dataset for agents in a realistic workplace
  setting.
\newblock In \emph{First Conference on Language Modeling (COLM)}, 2024.
\newblock URL \url{https://openreview.net/forum?id=4HNAwZFDcH}.

\bibitem[Su et~al.(2025)Su, Sun, Yoon, Yin, Yu, and
  Arik]{su2025learnbyinteract}
Hongjin Su, Ruoxi Sun, Jinsung Yoon, Pengcheng Yin, Tao Yu, and Sercan Arik.
\newblock Learn-by-interact: A data-centric framework for self-adaptive agents
  in realistic environments.
\newblock In \emph{International Conference on Learning Representations},
  volume 2025, pages 89044--89097, 2025.

\bibitem[Tang et~al.(2023)Tang, Deng, Lin, Han, Liang, Cao, and
  Sun]{tang2023toolalpaca}
Qiaoyu Tang, Ziliang Deng, Hongyu Lin, Xianpei Han, Qiao Liang, Boxi Cao, and
  Le~Sun.
\newblock Toolalpaca: Generalized tool learning for language models with 3000
  simulated cases.
\newblock \emph{arXiv preprint arXiv:2306.05301}, 2023.

\bibitem[{Team GLM} et~al.(2024){Team GLM}, Zeng, Xu, Wang, Zhang, Yin, Zhang,
  Rojas, Feng, Zhao, Lai, Yu, Wang, Sun, Zhang, Cheng, Gui, Tang, Zhang, Sun,
  Li, Zhao, Wu, Zhong, Liu, Huang, Zhang, Zheng, Lu, Duan, Zhang, Cao, Yang,
  Tam, Zhao, Liu, Xia, Zhang, Gu, Lv, Liu, Liu, Yang, Song, Zhang, An, Xu, Niu,
  Yang, Li, Bai, Dong, Qi, Wang, Yang, Du, Hou, and Wang]{glm2024chatglm}
{Team GLM}, Aohan Zeng, Bin Xu, Bowen Wang, Chenhui Zhang, Da~Yin, Dan Zhang,
  Diego Rojas, Guanyu Feng, Hanlin Zhao, Hanyu Lai, Hao Yu, Hongning Wang,
  Jiadai Sun, Jiajie Zhang, Jiale Cheng, Jiayi Gui, Jie Tang, Jing Zhang,
  Jingyu Sun, Juanzi Li, Lei Zhao, Lindong Wu, Lucen Zhong, Mingdao Liu, Minlie
  Huang, Peng Zhang, Qinkai Zheng, Rui Lu, Shuaiqi Duan, Shudan Zhang, Shulin
  Cao, Shuxun Yang, Weng~Lam Tam, Wenyi Zhao, Xiao Liu, Xiao Xia, Xiaohan
  Zhang, Xiaotao Gu, Xin Lv, Xinghan Liu, Xinyi Liu, Xinyue Yang, Xixuan Song,
  Xunkai Zhang, Yifan An, Yifan Xu, Yilin Niu, Yuantao Yang, Yueyan Li, Yushi
  Bai, Yuxiao Dong, Zehan Qi, Zhaoyu Wang, Zhen Yang, Zhengxiao Du, Zhenyu Hou,
  and Zihan Wang.
\newblock {ChatGLM}: A family of large language models from {GLM-130B} to
  {GLM-4} all tools, 2024.
\newblock URL \url{https://arxiv.org/abs/2406.12793}.

\bibitem[Trivedi et~al.(2024)Trivedi, Khot, Hartmann, Manku, Dong, Li, Gupta,
  Sabharwal, and Balasubramanian]{trivedi2024appworld}
Harsh Trivedi, Tushar Khot, Mareike Hartmann, Ruskin Manku, Vinty Dong, Edward
  Li, Shashank Gupta, Ashish Sabharwal, and Niranjan Balasubramanian.
\newblock Appworld: A controllable world of apps and people for benchmarking
  interactive coding agents.
\newblock In \emph{Proceedings of the 62nd Annual Meeting of the Association
  for Computational Linguistics (Volume 1: Long Papers)}, pages 16022--16076,
  2024.

\bibitem[Wang et~al.(2023{\natexlab{a}})Wang, Xie, Jiang, Mandlekar, Xiao, Zhu,
  Fan, and Anandkumar]{wang2023voyager}
Guanzhi Wang, Yuqi Xie, Yunfan Jiang, Ajay Mandlekar, Chaowei Xiao, Yuke Zhu,
  Linxi Fan, and Anima Anandkumar.
\newblock Voyager: An open-ended embodied agent with large language models,
  2023{\natexlab{a}}.
\newblock URL \url{https://arxiv.org/abs/2305.16291}.

\bibitem[Wang et~al.(2024{\natexlab{a}})Wang, Chen, Yuan, Zhang, Li, Peng, and
  Ji]{wang2024codeact}
Xingyao Wang, Yangyi Chen, Lifan Yuan, Yizhe Zhang, Yunzhu Li, Hao Peng, and
  Heng Ji.
\newblock Executable code actions elicit better llm agents.
\newblock \emph{arXiv preprint arXiv:2402.01030}, 2024{\natexlab{a}}.

\bibitem[Wang et~al.(2023{\natexlab{b}})Wang, Kordi, Mishra, Liu, Smith,
  Khashabi, and Hajishirzi]{wang2023selfinstruct}
Yizhong Wang, Yeganeh Kordi, Swaroop Mishra, Alisa Liu, Noah~A. Smith, Daniel
  Khashabi, and Hannaneh Hajishirzi.
\newblock Self-instruct: Aligning language models with self-generated
  instructions.
\newblock In \emph{Proceedings of ACL 2023 (Volume 1: Long Papers)},
  2023{\natexlab{b}}.
\newblock URL \url{https://arxiv.org/abs/2212.10560}.

\bibitem[Wang et~al.(2024{\natexlab{b}})Wang, Mao, Fried, and
  Neubig]{wang2024awm}
Zora~Zhiruo Wang, Jiayuan Mao, Daniel Fried, and Graham Neubig.
\newblock Agent workflow memory.
\newblock \emph{arXiv preprint arXiv:2409.07429}, 2024{\natexlab{b}}.

\bibitem[Xi et~al.(2024)Xi, Ding, Chen, Hong, Guo, Wang, Yang, Liao, Guo, He,
  Gao, Chen, Zheng, Zou, Gui, Zhang, Qiu, Huang, Wu, and Jiang]{xi2024agentgym}
Zhiheng Xi, Yiwen Ding, Wenxiang Chen, Boyang Hong, Honglin Guo, Junzhe Wang,
  Dingwen Yang, Chenyang Liao, Xin Guo, Wei He, Songyang Gao, Lu~Chen, Rui
  Zheng, Yicheng Zou, Tao Gui, Qi~Zhang, Xipeng Qiu, Xuanjing Huang, Zuxuan Wu,
  and Yu-Gang Jiang.
\newblock {AgentGym}: Evolving large language model-based agents across diverse
  environments, 2024.
\newblock URL \url{https://arxiv.org/abs/2406.04151}.

\bibitem[Yang et~al.(2025)Yang, Li, Yang, Zhang, Hui, Zheng, Yu, Gao, Huang,
  Lv, et~al.]{yang2025qwen3}
An~Yang, Anfeng Li, Baosong Yang, Beichen Zhang, Binyuan Hui, Bo~Zheng, Bowen
  Yu, Chang Gao, Chengen Huang, Chenxu Lv, et~al.
\newblock Qwen3 technical report.
\newblock \emph{arXiv preprint arXiv:2505.09388}, 2025.

\bibitem[Yao et~al.(2023{\natexlab{a}})Yao, Yu, Zhao, Shafran, Griffiths, Cao,
  and Narasimhan]{yao2023tot}
Shunyu Yao, Dian Yu, Jeffrey Zhao, Izhak Shafran, Thomas~L. Griffiths, Yuan
  Cao, and Karthik Narasimhan.
\newblock Tree of thoughts: Deliberate problem solving with large language
  models.
\newblock In \emph{Advances in Neural Information Processing Systems
  (NeurIPS)}, 2023{\natexlab{a}}.
\newblock URL \url{https://arxiv.org/abs/2305.10601}.

\bibitem[Yao et~al.(2023{\natexlab{b}})Yao, Zhao, Yu, Du, Shafran, Narasimhan,
  and Cao]{yao2023react}
Shunyu Yao, Jeffrey Zhao, Dian Yu, Nan Du, Izhak Shafran, Karthik Narasimhan,
  and Yuan Cao.
\newblock {ReAct}: Synergizing reasoning and acting in language models.
\newblock In \emph{International Conference on Learning Representations
  (ICLR)}, 2023{\natexlab{b}}.
\newblock URL \url{https://arxiv.org/abs/2210.03629}.

\bibitem[Yao et~al.(2024)Yao, Shinn, Razavi, and Narasimhan]{yao2024taubench}
Shunyu Yao, Noah Shinn, Pedram Razavi, and Karthik Narasimhan.
\newblock {$\tau$}-bench: A benchmark for tool-agent-user interaction in
  real-world domains.
\newblock \emph{arXiv preprint arXiv:2406.12045}, 2024.

\bibitem[Yu et~al.(2025)Yu, Peng, Vajipey, Cheng, Galley, Gao, and
  Yu]{yu2025exact}
Xiao Yu, Baolin Peng, Vineeth Vajipey, Hao Cheng, Michel Galley, Jianfeng Gao,
  and Zhou Yu.
\newblock Exact: Teaching ai agents to explore with reflective-mcts and
  exploratory learning.
\newblock In \emph{International Conference on Learning Representations},
  volume 2025, pages 65157--65184, 2025.

\bibitem[Yuan et~al.(2025)Yuan, Song, Chen, Tan, Shen, Ren, Li, and
  Yang]{yuan2025easytool}
Siyu Yuan, Kaitao Song, Jiangjie Chen, Xu~Tan, Yongliang Shen, Kan Ren,
  Dongsheng Li, and Deqing Yang.
\newblock Easytool: Enhancing llm-based agents with concise tool instruction.
\newblock In \emph{Proceedings of the 2025 Conference of the Nations of the
  Americas Chapter of the Association for Computational Linguistics: Human
  Language Technologies (Volume 1: Long Papers)}, pages 951--972, 2025.

\bibitem[Zelikman et~al.(2022)Zelikman, Wu, Mu, and Goodman]{zelikman2022star}
Eric Zelikman, Yuhuai Wu, Jesse Mu, and Noah~D. Goodman.
\newblock {STaR}: Bootstrapping reasoning with reasoning.
\newblock In \emph{Advances in Neural Information Processing Systems
  (NeurIPS)}, 2022.
\newblock URL \url{https://arxiv.org/abs/2203.14465}.

\bibitem[Zeng et~al.(2024)Zeng, Liu, Lu, Wang, Liu, Dong, and
  Tang]{zeng2024agenttuning}
Aohan Zeng, Mingdao Liu, Rui Lu, Bowen Wang, Xiao Liu, Yuxiao Dong, and Jie
  Tang.
\newblock Agenttuning: Enabling generalized agent abilities for llms.
\newblock In \emph{Findings of the Association for Computational Linguistics:
  ACL 2024}, pages 3053--3077, 2024.

\bibitem[Zeng et~al.(2025)Zeng, Liu, Wang, Li, Mi, Wang, Shang, Jiang, and
  Liu]{zeng2025toolacemt}
Xingshan Zeng, Weiwen Liu, Lingzhi Wang, Liangyou Li, Fei Mi, Yasheng Wang,
  Lifeng Shang, Xin Jiang, and Qun Liu.
\newblock {ToolACE-MT}: Non-autoregressive generation for agentic multi-turn
  interaction, 2025.
\newblock URL \url{https://arxiv.org/abs/2508.12685}.

\bibitem[Zhang et~al.(2025)Zhang, Lan, Zhu, Liu, Hoang, Kokane, Yao, Tan,
  Prabhakar, Chen, Liu, Feng, Awalgaonkar, Murthy, Hu, Chen, Xu, Niebles,
  Heinecke, Wang, Savarese, and Xiong]{zhang2025xlam}
Jianguo Zhang, Tian Lan, Ming Zhu, Zuxin Liu, Thai Hoang, Shirley Kokane,
  Weiran Yao, Juntao Tan, Akshara Prabhakar, Haolin Chen, Zhiwei Liu, Yihao
  Feng, Tulika Awalgaonkar, Rithesh Murthy, Eric Hu, Zeyuan Chen, Ran Xu,
  Juan~Carlos Niebles, Shelby Heinecke, Huan Wang, Silvio Savarese, and Caiming
  Xiong.
\newblock {xLAM}: A family of large action models to empower {AI} agent
  systems.
\newblock In \emph{Proceedings of the 2025 Conference of the North American
  Chapter of the Association for Computational Linguistics (NAACL)}, 2025.
\newblock URL \url{https://arxiv.org/abs/2409.03215}.

\bibitem[Zhang et~al.(2026)Zhang, Li, Xing, Apostolopoulos, Lee, and
  Zheng]{zhang2026gecko}
Zeyu Zhang, Guohao Li, Zhenchang Xing, Alexandros Apostolopoulos, Yu~Lin Lee,
  and Liang Zheng.
\newblock Gecko: A simulation environment with stateful feedback for refining
  agent tool calls, 2026.
\newblock URL \url{https://arxiv.org/abs/2602.19218}.

\bibitem[Zhao et~al.(2024)Zhao, Huang, Xu, Lin, Liu, and Huang]{zhao2024expel}
Andrew Zhao, Daniel Huang, Quentin Xu, Matthieu Lin, Yong-Jin Liu, and Gao
  Huang.
\newblock {ExpeL}: {LLM} agents are experiential learners.
\newblock In \emph{Proceedings of the AAAI Conference on Artificial
  Intelligence}, 2024.
\newblock URL \url{https://arxiv.org/abs/2308.10144}.

\bibitem[Zhao et~al.(2026)Zhao, Wen, Mao, Jiang, Hu, Wang, Zhang, and
  Wu]{zhao2026nextturn}
Yiwen Zhao, Zhihao Wen, Yuchen Mao, Mingxuan Jiang, Yihao Hu, Pan Wang, Xin
  Zhang, and Wei Wu.
\newblock Towards better agents for multi-turn user interaction: The next user
  turn is more than context.
\newblock \emph{arXiv preprint arXiv:2608.17499}, 2026.

\bibitem[Zhou et~al.(2023)Zhou, Yan, Shlapentokh-Rothman, Wang, and
  Wang]{zhou2023lats}
Andy Zhou, Kai Yan, Michal Shlapentokh-Rothman, Haohan Wang, and Yu-Xiong Wang.
\newblock Language agent tree search unifies reasoning acting and planning in
  language models, 2023.
\newblock URL \url{https://arxiv.org/abs/2310.04406}.

\bibitem[Zhou et~al.(2024)Zhou, Xu, Zhu, Zhou, Lo, Sridhar, Cheng, Ou, Bisk,
  Fried, Alon, and Neubig]{zhou2024webarena}
Shuyan Zhou, Frank~F. Xu, Hao Zhu, Xuhui Zhou, Robert Lo, Abishek Sridhar,
  Xianyi Cheng, Tianyue Ou, Yonatan Bisk, Daniel Fried, Uri Alon, and Graham
  Neubig.
\newblock {WebArena}: A realistic web environment for building autonomous
  agents.
\newblock In \emph{International Conference on Learning Representations
  (ICLR)}, 2024.
\newblock URL \url{https://arxiv.org/abs/2307.13854}.

\bibitem[Zhou et~al.(2026)Zhou, Uehara, Zhang, Zhou, Gu, Du, Xu, and
  Harada]{zhou2025toolgrad}
Zhongyi Zhou, Kohei Uehara, Haoyu Zhang, Jingtao Zhou, Lin Gu, Ruofei Du, Zheng
  Xu, and Tatsuya Harada.
\newblock {ToolGrad}: Efficient tool-use dataset generation with textual
  ``gradients''.
\newblock In \emph{Findings of ACL 2026}, 2026.
\newblock URL \url{https://arxiv.org/abs/2508.04086}.

\bibitem[Zhu et~al.(2026)Zhu, Fan, Wang, Wu, Zhou, and Huang]{zhu2026rsiagent}
Sibo Zhu, Shicheng Fan, Xinyue Wang, Wenyi Wu, Kun Zhou, and Biwei Huang.
\newblock {RSIAgent}: Autonomous exploration for recursive self-improvement in
  new environments, 2026.
\newblock URL \url{https://arxiv.org/abs/2609.15364}.

\bibitem[Zhuang et~al.(2023)Zhuang, Yu, Wang, Sun, and Zhang]{zhuang2023toolqa}
Yuchen Zhuang, Yue Yu, Kuan Wang, Haotian Sun, and Chao Zhang.
\newblock {ToolQA}: A dataset for {LLM} question answering with external tools.
\newblock In \emph{Advances in Neural Information Processing Systems
  (NeurIPS)}, 2023.
\newblock URL \url{https://arxiv.org/abs/2306.13304}.

\end{thebibliography}
\bibliographystyle{plainnat}

\onecolumn
\appendix
\small
\raggedbottom
\FloatBarrier
\section{Experimental details}
\label{app:setup-details}
\stitle{Benchmark handling.}
Accuracy uses the official BFCL scorer on all tasks and counts context exhaustion and residual transport failures as failures. Executions interrupted by transport errors are rerun under fixed rules, and only the final execution enters accuracy and cost.

\stitle{Diagnostics.}
$R$, $P$, and $B$ use the fixed requirement annotations and scoring rules without refitting. Table~\ref{tab:metrics-story} and Figure~\ref{fig:metrics-transitions} pool the disjoint Train400 and Test400 tasks; AUROC uses tasks with defined Plain diagnostics, paired changes use tasks with defined diagnostics in both executions, and undefined scores are excluded, never set to zero. Figure~\ref{fig:metrics-transitions} uses untrained checkpoints for all eight models, averaging within-run transition means over three repeats each for Qwen3.8-27B and GLM-5.3 and using one Plain/LTP pair for each other model.
For Table~\ref{tab:metrics-story}, the valid sample sizes ($n$) out of 800 tasks per model are Qwen2.5-7B: 794; Qwen3-8B: 795; Qwen3.5-9B: 797; Qwen3-14B: 798; Qwen3.5-397B: 798; DeepSeek-V4-Flash: 798. These counts apply to all three diagnostics.

\stitle{Baselines.}
All baselines are adapted to the BFCL interface. Reflexion uses public feedback and a fixed second trial, is scored on that trial, and is charged for the first trial and the reflection; RSIAgent is charged for its verifier calls. Gecko/GATS covers all 800 tasks; five tasks with incomplete usage records were rerun, replacing their outcomes, diagnostics, and costs together. Inference seeds in Table~\ref{tab:inference800}: Plain 116931, Gecko/GATS 116933, LTP run 116939.

\stitle{Cost accounting.}
Token cost counts the input and output tokens of the evaluated execution and its method-specific helper calls, plus preparation counted once and shared by the tasks that use it. Failed tasks stay in the denominator, and their costs count. Gecko's helper tokens are attributed to the trajectories; helper tokens from superseded attempts are excluded.

\stitle{Training.}
The loss is token-normalized over assistant decisions, with the preceding history as context. The training seed is 42 for the Qwen2.5-7B, Qwen3-8B, and Qwen3.5-9B students in Table~\ref{tab:sft127} and for the fine-tuned Qwen3-8B checkpoints in Appendix~\ref{app:cases}; the Qwen3-14B student uses seed 5.

\FloatBarrier
\section{Case studies from recorded trajectories}
\label{app:cases}
\subsection{Trace examples for the execution diagnostics}
\label{app:metric-trace-cases}
Three completed-task comparisons show how the diagnostics relate to recorded calls (Table~\ref{tab:metric-trace-cases}). The budget and directory examples use one Qwen3-8B checkpoint trained on LTP trajectories with the bank context removed (training seed 42, learning rate $5\times10^{-6}$, epoch 2); the ticket example compares the selected DeepSeek-V4-Flash-0731 LTP run with the adapted Reflexion run. The cases were selected after inspecting outcomes.

\begin{table}[!htb]
\centering\small
\caption{\textbf{Diagnostic changes have concrete causes in the recorded calls.} Three BFCL tasks; each tuple is $(R,P,B)$, arrows follow the execution order in the second column, and IDs omit the \texttt{multi\_turn\_} prefix.}
\label{tab:metric-trace-cases}
\setlength{\tabcolsep}{4pt}
\begin{tabular}{@{}lll@{}}
\toprule
Task & Executions and official outcomes & $(R,P,B)$ \\
\midrule
\texttt{miss\_param\_150} & Plain fail $\to$ LTP pass & $(1,.5,.5)\to(1,1,1)$ \\
\texttt{miss\_param\_22} & Plain pass $\to$ LTP fail & $(.75,.75,.75)\to(.5,.5,.5)$ \\
\texttt{long\_context\_23} & Reflexion fail $\to$ LTP pass & $(\frac23,\frac23,\frac23)\to(1,1,1)$ \\
\bottomrule
\end{tabular}
\end{table}

In \texttt{miss\_param\_150}, both executions obtain the flight fare and the exchange rate, but Plain sets the budget to the unconverted 15,400 while LTP converts 15,400 GBP to 22,000 USD and passes that value; $R$ stays 1 and $P$ and $B$ rise from .5 to 1.

In \texttt{miss\_param\_22}, both copy the file, but LTP then changes into the archive directory, where the comparison fails on a missing file; the comparison requirement loses its witness and all three scores fall by .25 with a pass-to-fail outcome.

In \texttt{long\_context\_23}, Reflexion reads ticket 7423 but never resolves it, whereas LTP calls \texttt{resolve\_ticket}; coverage is $2/3$ versus 1.

\subsection{Selected LTP run and alternative inference trajectories}
\label{app:inference-trace-cases}
Table~\ref{tab:inference-trace-cases} lists three tasks on which the selected LTP run succeeds and the named alternatives fail; the baseline adaptations use a different inference seed, and EasyTool comes from its separate run. In \texttt{miss\_param\_109}, LTP waits for the missing quantity before placing the order and then inspects and cancels the returned order, whereas P2P places an unsolicited 10-share order first and Reflexion retrieves an old order instead of buying; P2P's later correction still earns $(1,1,1)$, so coverage does not record every premature action.

In \texttt{miss\_func\_189}, LTP books the flight once \texttt{book\_flight} becomes available, whereas P2P and EasyTool call \texttt{verify\_traveler\_information} with invented passport data and never book.

In \texttt{base\_54}, both LTP and P2P receive an unlocked-doors error from \texttt{startEngine}; LTP locks the doors, presses the brake fully, and starts the engine, while P2P moves on with the engine stopped. LTP succeeds through recovery from observed errors; its trajectory is not error-free.

\begin{table}[!htb]
\centering\small
\caption{\textbf{Where LTP succeeds and the named alternatives fail.} Three BFCL tasks with $(R,P,B)$ for every displayed execution; IDs omit \texttt{multi\_turn\_}.}
\label{tab:inference-trace-cases}
\setlength{\tabcolsep}{4pt}
\begin{tabular}{@{}lll@{}}
\toprule
Task & LTP $(R,P,B)$ & Comparator $(R,P,B)$ \\
\midrule
\texttt{miss\_param\_109} & $(1,1,1)$ & P2P $(1,1,1)$; Reflexion $(\frac47,\frac47,\frac37)$ \\
\texttt{miss\_func\_189} & $(1,1,1)$ & P2P and EasyTool $(.75,.75,.75)$ \\
\texttt{base\_54} & $(.75,.75,.5)$ & P2P $(.5,.5,.25)$ \\
\bottomrule
\end{tabular}
\end{table}

\subsection{Matched SFT trajectory comparisons}
\label{app:sft-trace-cases}
Three BFCL cases compare Qwen3-8B checkpoints trained on Plain data and on LTP data with the bank context retained, at the same training seed (42), learning rate ($2\times10^{-6}$), and epoch (3), both evaluated with Plain inference and no retrieved bank; the untrained model is a reference from an earlier inference configuration. The Retained SFT checkpoint succeeds in all three cases, and the Plain SFT checkpoint and the untrained model fail (Table~\ref{tab:sft-trace-cases}). The cases were selected after inspecting outcomes.

\begin{table}[ht]
\centering\small
\caption{\textbf{Retained SFT completes the steps the other students skip.} Correctness and diagnostics for the same three tasks under Plain inference. The case IDs are \texttt{multi\_turn\_miss\_param\_150}, \texttt{multi\_turn\_long\_context\_24}, and \texttt{multi\_turn\_long\_context\_86}.}
\label{tab:sft-trace-cases}
\begin{tabular}{llcccc}
\toprule
Case & Training data & Correct & $R$ & $P$ & $B$ \\
\midrule
Budget & Untrained & No & .500 & .500 & .500 \\
 & Plain SFT & No & .500 & .500 & .000 \\
 & Retained SFT & Yes & 1.000 & 1.000 & 1.000 \\
\midrule
Move and rename & Untrained & No & .667 & .500 & .500 \\
 & Plain SFT & No & .833 & .500 & .333 \\
 & Retained SFT & Yes & .833 & .833 & .667 \\
\midrule
Vehicle workflow & Untrained & No & .857 & .857 & .714 \\
 & Plain SFT & No & .714 & .714 & .714 \\
 & Retained SFT & Yes & 1.000 & 1.000 & 1.000 \\
\bottomrule
\end{tabular}
\end{table}

In the budget case, the Retained SFT model converts 15,400 GBP to 22,000 USD and passes the converted value to the budget setter, after recovering from an initial invalid airport-code guess, whereas the Plain SFT and untrained models compute the conversion but never set the budget.

In the move-and-rename case, the Retained SFT model moves the file, enters the archive, and renames it, whereas the Plain SFT model repeats the move and leaves the new name unapplied, and the untrained model puts a path in the destination and is rejected.

In the vehicle case, the Retained SFT model engages the parking brake, corrects a half brake press to a full one after the returned error, starts the engine, and completes the navigation and tweet steps, whereas the Plain SFT model also recovers the engine start but never engages the parking brake or posts the tweet, and the untrained model abandons the engine recovery after the first brake error and navigates to a substituted destination. These are learned behaviors under Plain inference; no evidence is retrieved at test time.

\FloatBarrier

\FloatBarrier
\section{Reference results with the official-API DeepSeek-V4.1-Flash}
\label{app:v41}
The main text reports every DeepSeek result with the 0731 API version. Earlier runs used the official-API DeepSeek-V4.1-Flash; Table~\ref{tab:v41} keeps its Test400 diagnostics for reference, together with the Plain-inference AUROC on all 800 tasks for the two Figure~\ref{fig:metrics-transitions} models that Table~\ref{tab:metrics-story} omits (three Plain runs each). With DeepSeek-V4.1-Flash, LTP solved 31 of 49 ACEBench Agent tasks against 27 for Plain, and 56 of 78 ToolTalk conversations against 53.

\begin{table}[!htb]
\centering\small
\caption{Reference diagnostics not used in the main text. DeepSeek-V4.1-Flash results use Test400; Qwen3.8-27B and GLM-5.3 AUROCs use all 800 tasks, averaged over three runs.}
\label{tab:v41}
\setlength{\tabcolsep}{5pt}
\begin{tabular}{@{}llrrr@{}}
\toprule
Model & Quantity & $R$ & $P$ & $B$ \\
\midrule
DeepSeek-V4.1-Flash & Plain AUROC (397 valid tasks) & 0.704 & 0.715 & 0.740 \\
 & Mean $\Delta$ on failure $\to$ success tasks & $+0.105$ & $+0.145$ & $+0.068$ \\
 & Mean $\Delta$ on success $\to$ failure tasks & $-0.149$ & $-0.151$ & $-0.096$ \\
Qwen3.8-27B & Plain AUROC (795, 796, 791 valid tasks) & 0.717 & 0.734 & 0.665 \\
GLM-5.3 & Plain AUROC (798 valid tasks per run) & 0.660 & 0.691 & 0.619 \\
\bottomrule
\end{tabular}
\end{table}

\clearpage
\section{Additional evaluations and controlled experiments}
\label{app:or-additional}
\subsection{Repeated evaluation and task sets}
\label{app:or-repeats}
The additional DeepSeek-V4-Flash experiments use a newly constructed bank with 80 probe episodes and 352 records. Their results are compared within that batch, separately from the original runs in Table~\ref{tab:inference800}. The additional task set contains 400 tasks written after submission and sealed before any run. Inference seeds for the three new DeepSeek pairs are fixed in advance; the Qwen3-8B repeated evaluation rebuilds the bank for every seed. Unless otherwise specified, the reported 95\% intervals bootstrap task families while pairing seeds. GLM-5.3 instead uses a Student's $t$ interval over its three retained seed-level gains.

\begin{table}[!htbp]
\centering\small
\caption{\textbf{Repeated runs across three models.} Solved tasks are Plain $\to$ LTP within each pair. Gains and intervals are in percentage points. The Qwen3-8B evaluation rebuilds its bank for each seed.}
\label{tab:or-repeats}
\setlength{\tabcolsep}{5pt}
\begin{tabular}{@{}llrrrll@{}}
\toprule
Model & Tasks & Pair 1 & Pair 2 & Pair 3 & Mean gain & 95\% interval\\
\midrule
DeepSeek-V4-Flash & BFCL800 & $301\to366$ & $290\to343$ & $298\to331$ & +6.29 & [4.17,8.42]\\
DeepSeek-V4-Flash & 400 new & $335\to346$ & $327\to363$ & $336\to363$ & +6.17 & [1.83,11.58]\\
Qwen3-8B & 400 new & $145\to169$ & $144\to171$ & $143\to170$ & +6.50 & [1.67,12.50]\\
GLM-5.3 & BFCL800 & $431\to456$ & $433\to469$ & $423\to457$ & +3.96 & [2.14,5.78]\\
\bottomrule
\end{tabular}
\end{table}

Each DeepSeek-V4-Flash BFCL pair improves, with exact McNemar $p$ values of $1.5\times10^{-6}$, $7.5\times10^{-5}$, and $.020$, respectively. The gains are positive in all twelve category--seed combinations. Category-level mean gains are 6.8 points on base tasks, 7.8 on long-context tasks, 4.8 on missing-function tasks, and 5.7 on missing-parameter tasks. GLM-5.3 improves in all three retained pairs; its Plain mean is 53.63\%, compared with 37.04\% for DeepSeek-V4-Flash in the additional batch.

\subsection{Evidence representation controls}
\label{app:or-content}
Every non-Plain condition in Table~\ref{tab:or-representations} uses the same probing run, bank, and selector within its model's experiment. Only the content delivered to the agent changes. Matching trims each form to the shortest representation at every step, token for token, and renders it in a common plain format. The probing budget is therefore shared. DeepSeek-V4-Flash is evaluated on BFCL800 and Qwen3-8B on the 400 new tasks. The means use three seeds, except the DeepSeek documentation and instruction rows, which use two. The rebuilt-bank Qwen repeated evaluation in Table~\ref{tab:or-repeats} is distinct from this fixed-bank representation comparison.

\begin{table}[!htbp]
\centering\small
\caption{\textbf{Token-matched evidence representations.} Accuracy (\%). Dashes denote representations not reported for that model. Documentation repeats information already available to the agent. Faithful summaries and usage instructions rewrite the same observations after probing; short probing notes are written during probing.}
\label{tab:or-representations}
\setlength{\tabcolsep}{9pt}
\begin{tabular}{@{}lrr@{}}
\toprule
Delivered content & DeepSeek-V4-Flash & Qwen3-8B\\
\midrule
Plain (no added evidence) & 37.04 & 36.42\\
Raw records & 40.00 & 43.00\\
Calls only, without responses & 36.42 & 38.92\\
Other tools' records & 38.25 & 39.50\\
Documentation & 38.69 & 40.00\\
Usage instructions & 40.56 & ---\\
Faithful summary & 40.17 & ---\\
Short probing notes & --- & 35.67\\
\bottomrule
\end{tabular}
\end{table}

On DeepSeek-V4-Flash, Raw minus Calls-only is +3.58 points ($[1.38,6.00]$); under the original 1,800-character cap instead of exact token matching, the difference is +2.17 points ($[0.21,4.17]$). On Qwen3-8B, the token-matched difference is +4.1 points ($[0.1,8.7]$). Other tools' records differ from Plain by +1.21 points ($[-0.87,3.33]$) on DeepSeek-V4-Flash. On matched seeds, repeated documentation is 2.1 points below Raw for DeepSeek-V4-Flash and 3.0 points below for Qwen3-8B.

Raw minus faithful summaries is $-0.17$ points ($[-2.04,1.71]$), and Raw minus usage instructions is +0.25 points ($[-2.38,2.81]$), using the matched seeds for each contrast. These intervals allow similar performance for faithful rewrites. Both rewrites require an additional model pass over the bank; raw records preserve the observations directly. On Qwen3-8B, Raw exceeds short probing notes by 7.3 points ($[2.3,14.2]$).

\paragraph{Deployed format versus the matched renderer.}
LTP with its deployed format and full budget reaches 43.33\% on the three DeepSeek-V4-Flash seeds. Its differences from the token-matched Calls-only and Other-tools conditions are +6.92 points ($[4.71,9.21]$) and +5.08 points ($[2.71,7.38]$). Under a shared 1,800-character cap, deployed LTP exceeds repeated documentation by 3.38 points ($[0.94,5.75]$), instructions by 5.38 points ($[2.88,7.88]$), and faithful summaries by 4.54 points ($[2.42,6.62]$). These deployed-versus-control comparisons also involve the rendering format and must be distinguished from the content-only, exact-token contrasts above.

\subsection{Eleven probing, storage, and retrieval variants}
\label{app:or-ablations}
Each variant changes one decision on the same 800 BFCL tasks with DeepSeek-V4-Flash and two paired seeds. Matched Plain and full LTP score 36.94\% and 44.31\%, respectively. Table~\ref{tab:or-ablations} reports the mean accuracy, difference from full LTP, its 95\% interval, and the fraction of LTP's gain over Plain that remains. The two half-budget variants each use 40 of the original 80 probe episodes.

\begin{table}[!htbp]
\centering\small
\caption{\textbf{Complete design ablations.} Accuracy in percent; differences and intervals in percentage points. Gain kept is relative to the matched full-LTP improvement over Plain.}
\label{tab:or-ablations}
\setlength{\tabcolsep}{5pt}
\begin{tabular}{@{}llrrlr@{}}
\toprule
Decision & Variant & Accuracy & $\Delta$ vs. full & 95\% interval & Gain kept\\
\midrule
 & Full LTP & 44.31 & --- & --- & 100\%\\
Tools & Evidence for half the functions & 41.56 & $-2.75$ & [$-4.94,-0.56$] & 63\%\\
Tools & Random functions, no goals & 43.06 & $-1.25$ & [$-3.44,+1.00$] & 83\%\\
Tools & Half budget, all eight domains & 41.00 & $-3.31$ & [$-5.69,-0.94$] & 55\%\\
Tools & Half budget, four of eight domains & 38.00 & $-6.31$ & [$-8.75,-3.88$] & 14\%\\
Responses & One record per function & 41.38 & $-2.94$ & [$-5.13,-0.75$] & 60\%\\
Retrieval & Random order, no ranking & 41.69 & $-2.63$ & [$-5.06,-0.19$] & 64\%\\
Retrieval & $k=1$ record & 41.31 & $-3.00$ & [$-5.25,-0.81$] & 59\%\\
Retrieval & $k=5$ records & 42.38 & $-1.94$ & [$-4.06,+0.25$] & 74\%\\
Retrieval & $m=600$ characters & 39.25 & $-5.06$ & [$-7.31,-2.81$] & 31\%\\
Retrieval & $m=3{,}600$ characters & 43.25 & $-1.06$ & [$-3.13,+1.06$] & 86\%\\
Retrieval & No visible-tool filter & 43.00 & $-1.31$ & [$-3.56,+0.94$] & 82\%\\
\bottomrule
\end{tabular}
\end{table}

Distributing 40 episodes over all eight domains exceeds the four-domain allocation by 3.00 points ($[0.38,5.63]$), and exceeds Plain by 4.06 points ($[1.56,6.63]$). Broad tool coverage matters even at a fixed episode budget. Keeping only one response per function also reduces the gain. Random function probing retains most of the gain, with a difference from full LTP whose interval includes zero. Random retrieval order, one-record delivery, and the 600-character cap reduce accuracy, whereas the larger tested budgets provide no improvement in their point estimates. All variants remain above Plain in point estimate. Error responses are already excluded by the delivery rule in Section~\ref{sec:ltp}.

\subsection{Additional baseline comparisons}
\label{app:or-baselines}
The original Table~\ref{tab:inference800} shows accuracy differences from EasyTool and Gecko/GATS that are not significant under exact McNemar tests ($p=.47$ and $.54$). LTP exceeds RSIAgent by 5.38 points ($[0.88,10.00]$, $p=.005$), and exceeds Reflexion by 13.9 points. The two new paired-seed comparisons against EasyTool give a 95\% interval of $[-3.08,3.86]$ points for LTP minus EasyTool.

\begin{table}[!htbp]
\centering\small
\caption{\textbf{Qwen3-8B baselines on 400 newly written tasks.} Means over three paired seeds. LTP records denotes the token-matched Raw condition in Table~\ref{tab:or-representations}. Differences and intervals are relative to Plain, in percentage points.}
\label{tab:or-baselines}
\setlength{\tabcolsep}{9pt}
\begin{tabular}{@{}lrrl@{}}
\toprule
Method & Accuracy (\%) & $\Delta$ vs. Plain & 95\% interval\\
\midrule
Plain & 36.42 & --- & ---\\
Play2Prompt & 36.25 & $-0.17$ & [$-4.50,+3.50$]\\
EasyTool & 42.00 & $+5.58$ & [$-1.25,14.08$]\\
LTP records & 43.00 & $+6.58$ & [1.75,12.83]\\
\bottomrule
\end{tabular}
\end{table}

LTP records exceed Play2Prompt by 6.75 points ($[2.00,12.17]$) and have a one-point higher accuracy estimate than EasyTool. This fixed-bank comparison has different Plain/Raw outcomes from the rebuilt-bank Qwen3-8B repetition in Table~\ref{tab:or-repeats}; they are separate experiments on the same new task set.

For ToolAtlas, we implement the method from its paper \citep{fang2026toolatlas}, with a preparation cap of 8.18M tokens, equal to the original LTP bank cost. On Qwen3-8B and BFCL800, its mean accuracy is 11.56\%, compared with 13.56\% for Plain and 16.13\% for LTP. This comparison is separate from the 400-task baseline table.

\subsection{Preparation, reuse, and state sensitivity}
\label{app:or-cost}
The two bank builds cost 8.18M and 5.14M tokens, respectively, equivalent to approximately 47--61 Plain task executions in their respective BFCL batches. Preparation uses ten probe episodes per domain. In the additional three-seed batch, mean LTP execution tokens are 3.1\% below Plain; in the original run they are 1.7\% above Plain. Across the additional batch's 2,400 task executions, the totals are 256.8M for LTP including the shared bank and 259.6M for Plain. These 2,400 executions repeat the same BFCL800 tasks across three seeds.

\begin{table}[!htbp]
\centering\small
\caption{\textbf{Token costs for the two additional EasyTool comparisons.} Execution totals per 800 tasks, in millions of tokens. The LTP bank costs 5.14M tokens once.}
\label{tab:or-cost}
\setlength{\tabcolsep}{10pt}
\begin{tabular}{@{}lrr@{}}
\toprule
Method & First paired run & Second paired run\\
\midrule
EasyTool execution & 126.5 & 127.4\\
LTP execution & 84.1 & 84.5\\
\bottomrule
\end{tabular}
\end{table}

Under the reported execution rates and reuse of one bank, total LTP tokens fall below EasyTool within 100 tasks, and below Gecko/GATS from the 15th task. These break-even counts compare against those baselines, not Plain, and project the observed rates to reuse at different task counts. In the original comparison, the incremental token cost per additional solved task is 0.16M for LTP, versus 2.1M--8.6M for baselines that improve on Plain.

The delivered evidence remains capped at three records and 1,800 characters per decision. Removing the visible-tool filter expands the candidate pool to the whole 352-record bank across eight domains, yielding 43.00\% versus 44.31\% with the filter ($-1.31$ points, $[-3.56,0.94]$). The original AppWorld evaluation spans 457 APIs across nine apps, where success rises from 13.33\% to 20.85\%. These results characterize a larger candidate pool and one larger tool environment; they do not measure a preparation-cost scaling curve.

For state sensitivity, all identifiers in the bank are replaced with new names, and Qwen3-8B is evaluated on 200 file-system tasks over two seeds. Accuracy changes by $-0.50$ points ($[-2.25,1.00]$). The bank therefore need not supply matching live identifiers to retain the observed gain. A domain whose tool behavior changes can be probed again using the same ten-episode allocation; this is a maintenance procedure rather than a measured behavior-change result.

\subsection{SFT with retrieved evidence removed}
\label{app:or-sft}
We remove retrieved evidence from the training context of LTP trajectories for Qwen2.5-7B, Qwen3-8B, and Qwen3.5-9B. The comparison covers 24 matched configurations across two learning rates and four epochs. Evaluation uses Plain inference without a bank. Removing the evidence changes mean accuracy by +0.4 percentage points; accuracy is higher in 13 configurations and lower in 11, and 21 of the 24 reported intervals include zero. Table~\ref{tab:or-sft} summarizes the reported control.

\begin{table}[!htbp]
\centering\small
\caption{\textbf{Removing evidence from SFT context.} Changes are evidence-removed minus evidence-retained training, evaluated with Plain inference. The teacher's action targets are retained.}
\label{tab:or-sft}
\setlength{\tabcolsep}{8pt}
\begin{tabular}{@{}lr@{}}
\toprule
Quantity & Result\\
\midrule
Student models & 3\\
Matched configurations & 24\\
Learning rates / epochs & 2 / 4\\
Mean accuracy change (pp) & +0.4\\
Configurations improving / declining & 13 / 11\\
Intervals including zero & 21 / 24\\
\bottomrule
\end{tabular}
\end{table}

Both original training arms use the same 127 tasks and teacher, and evidence is loss-masked when retained. The removal control shows a small average effect from that context while preserving the teacher's action trajectories. The ToolGrad and ToolACE-MT comparisons in Figure~\ref{fig:sft8b-baselines} draw successful trajectories from the same 400 training tasks and teacher, but are not limited to the same 127 jointly successful tasks; they use 126 and 127 trajectories, respectively. These experiments address trajectory construction and evidence context, without reporting a matched training-token-budget comparison.

\subsection{Diagnostics when success does not change}
\label{app:or-diagnostics}
A further check on the original main run examines the 618 tasks whose success outcome stays unchanged between Plain and LTP. Mean argument-constraint coverage changes by at most 0.002 on these tasks, compared with +0.165 on rescued tasks and $-0.225$ on harmed tasks in Section~\ref{sec:inference}. The diagnostics provide graded trace-level measurements even when success is binary. Their association with outcome changes and the separate content interventions give complementary evidence: one localizes fulfilled requirements, while the other changes what evidence the agent receives.

\end{document}